\documentclass[ruled,vlined,linesnumbered]{article}

\usepackage{microtype}
\usepackage{graphicx}
\usepackage{subcaption}
\usepackage{booktabs} 
\usepackage{hyperref}

\usepackage[ruled,vlined,algo2e]{algorithm2e}

\usepackage[preprint]{icml2026}

\usepackage{amsmath}
\usepackage{amssymb}
\usepackage{mathtools}
\usepackage{amsthm}

\usepackage{amssymb}
\usepackage{pifont}

\usepackage{xcolor}

\usepackage[table]{xcolor}

\definecolor{checkgreen}{RGB}{76,175,80}   
\definecolor{crossorange}{RGB}{245,124,0}  
\definecolor{headergray}{RGB}{230,230,240}
\definecolor{rowgray}{gray}{0.92}

\newcommand{\cmark}{\textcolor{checkgreen}{\ding{51}}}
\newcommand{\xmark}{\textcolor{crossorange}{\ding{55}}}

\usepackage[most]{tcolorbox}
\newtcolorbox{methodbox}[1]{
  colback=white,
  colframe=black,
  coltitle=white,
  fonttitle=\bfseries,
  title=#1,
  boxrule=0.8pt,
  arc=0mm,
  left=6pt,
  right=6pt,
  top=6pt,
  bottom=6pt
}

\usepackage{etoc}

\newcommand{\bluecomment}[1]{\textcolor{blue}{\textbf{/* #1 */}}}

\usepackage{listings}
\usepackage{xcolor}
\usepackage{tcolorbox}

\tcbuselibrary{listings,skins,breakable}

\usepackage{enumitem}

\usepackage{mdframed}

\newmdenv[
  leftline=true,
  rightline=false,
  topline=false,
  bottomline=false,
  linecolor=gray,
  linewidth=2pt,
  backgroundcolor=gray!8,
  innerleftmargin=8pt,
  innerrightmargin=8pt,
  innertopmargin=6pt,
  innerbottommargin=6pt
]{insightbox}

\usepackage{placeins}

\usepackage[capitalize,noabbrev]{cleveref}

\theoremstyle{plain}

\theoremstyle{definition}

\theoremstyle{remark}

\usepackage[textsize=tiny]{todonotes}

\icmltitlerunning{Who Deserves the Reward? SHARP: Shapley Credit-based Optimization for Multi-Agent System}

\begin{document}

\twocolumn[
\icmltitle{Who Deserves the Reward? \\ SHARP: Shapley Credit-based Optimization for Multi-Agent System}


  \icmlsetsymbol{equal}{*}

  \begin{icmlauthorlist}
    \icmlauthor{Yanming Li}{equal,yyy,sch3}
    \icmlauthor{Xuelin Zhang}{equal,yyy,sch3}
    \icmlauthor{Wenjie Lu}{equal,yyy}
    
    \icmlauthor{Ziye Tang}{sch4}
    \icmlauthor{Maodong Wu}{sch3}
    \icmlauthor{Haotian Luo}{sch3}
    \icmlauthor{Tongtong Wu}{sch5}
    \icmlauthor{Zijie Peng}{sch3}
    \icmlauthor{Hongze Mi}{yyy,sch2}
    \icmlauthor{Yibo Feng}{sch1}
    \icmlauthor{Naiqiang Tan}{yyy}
    \icmlauthor{Chao Huang}{sch3}
    \icmlauthor{Lian Peng}{sch6}
    \icmlauthor{Li Shen}{sch3}
  \end{icmlauthorlist}

  \icmlaffiliation{yyy}{Didichuxing Co. Ltd}
  \icmlaffiliation{sch1}{The Chinese University of Hong Kong, Shenzhen}
  \icmlaffiliation{sch2}{Tianjin University}
  \icmlaffiliation{sch3}{Sun Yat-sen University}
  \icmlaffiliation{sch4}{Southeast University}
  \icmlaffiliation{sch5}{Monash University}
  \icmlaffiliation{sch6}{Huazhong Agricultural University}

  \icmlcorrespondingauthor{Naiqiang Tan}{tannaiqiang@didiglobal.com}
  \icmlcorrespondingauthor{Li Shen}{shenli6@mail.sysu.edu.cn}

  \icmlkeywords{Machine Learning, ICML}

  \vskip 0.3in
]

\newcommand{\ls}[1]{{\color{red}{[LS: #1]}}}



\printAffiliationsAndNotice{}  

\begin{abstract}
Integrating Large Language Models (LLMs) with external tools via multi-agent systems offers a promising new paradigm for decomposing and solving complex problems. However, training these systems remains notoriously difficult due to the credit assignment challenge, as it is often unclear which specific functional agent is responsible for the success or failure of decision trajectories. Existing methods typically rely on sparse or globally broadcast rewards, failing to capture individual contributions and leading to inefficient reinforcement learning. 
To address these limitations, we introduce the Shapley-based Hierarchical Attribution for Reinforcement Policy (SHARP), a novel framework for optimizing multi-agent reinforcement learning via precise credit attribution. SHARP effectively stabilizes training by normalizing agent-specific advantages across trajectory groups, primarily through a decomposed reward mechanism comprising a global broadcast-accuracy reward, a Shapley-based marginal-credit reward for each agent, and a tool-process reward to improve execution efficiency. Extensive experiments across various real-world benchmarks demonstrate that SHARP significantly outperforms recent state-of-the-art baselines, achieving average match improvements of 23.66\% and 14.05\% over single-agent and multi-agent approaches, respectively. 

\end{abstract}

\section{Introduction}
\begin{figure}[t]
\centering
\includegraphics[width=\columnwidth]{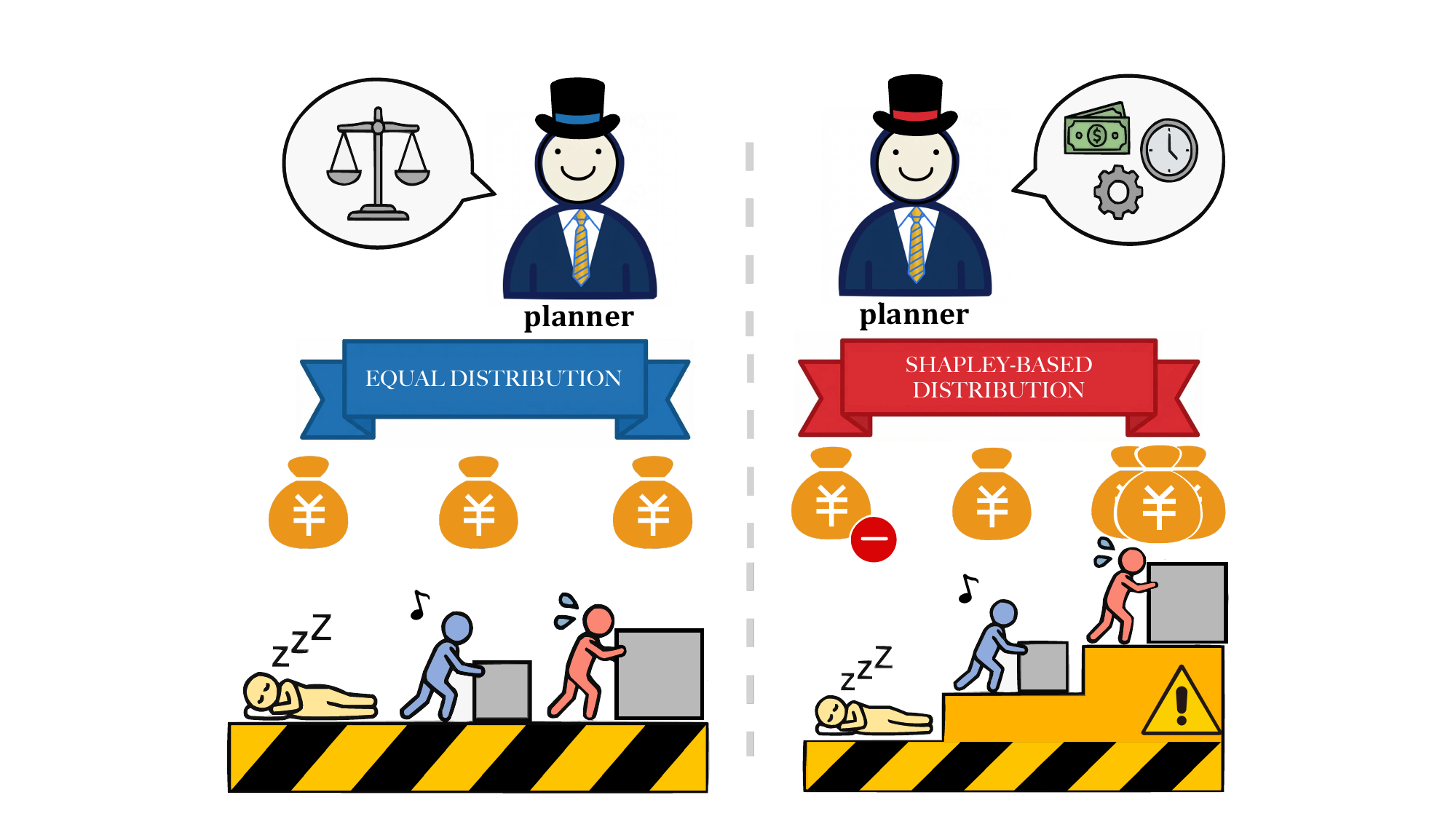}
\caption{Existing credit assignment policy for all agents (left) and the precise strategy of SHARP for each individual agent (right).}
\label{fig:intro_head}
\end{figure}

The evolution of Large Language Models (LLMs) has enabled a fundamental shift from static knowledge retrieval to dynamic and tool-augmented interactions in complex real-world scenarios \cite{lewis2020retrieval,wang2024empowering}. Integrating LLMs with external tools via multi-agent systems (MAS) offers a promising new paradigm for decomposing and solving complex problems that require reasoning, planning, and execution \cite{zhang2025cu,ferrag2025llm}. 
Unlike single-agent architectures that frequently struggle with context overflow and noisy tool feedback, MAS frameworks employ a collaborative structure in which a planner agent manages task decomposition, while specialized worker agents execute subtasks \cite{li2025webthinker}. 
Despite these advantages, training such systems remains challenging due to the inherent complexity of credit assignment. In MAS with collaborative settings where a planner defines strategies and workers execute subtasks, it is inherently unclear whether the success or failure of a trajectory should be attributed to high-level planning or to specific execution. 

Current research has diverged into preference-based Reinforcement Learning (RL) \cite{schulman2017proximal} and specialized Multi-Agent Reinforcement Learning (MARL) \cite{motwani2024malt,zhao2025one} architectures. On one hand, value-free optimization methods such as Direct Preference Optimization (DPO) \cite{rafailov2023direct} and Group Relative Policy Optimization (GRPO) \cite{shao2024deepseekmath} have been developed to eliminate the instability of critic estimation in uncertain environments. On the other hand, recent MARL frameworks such as Multi-Agent Tool-augmented Policy Optimization (MATPO) \cite{mo2025multi} and AceSearcher \cite{xu2025acesearcher} have introduced planner and worker hierarchies to better manage task complexity. Despite these advancements, both categories primarily rely on sparse and globally broadcast rewards that treat multiple agents with different roles as a monolithic entity. Such reliance on aggregate outcomes hinders the isolation of individual agents' marginal contributions, thereby yielding inefficient policy updates that obscure high-quality individual actions amid overall team performance.

To address these limitations in multi-agent credit assignment, we propose the Shapley-based Hierarchical Attribution for Reinforcement Policy (SHARP), a novel framework that stabilizes multi-agent training through precise, fine-grained credit attribution. Moving beyond monolithic broadcast signals, SHARP is designed with a tripartite decomposed reward principle which incorporates 1) a global broadcast-accuracy term for task alignment, 2) a marginal-credit reward based on Shapley values to quantify individual impact, and 3) a tool-process reward that ensures execution validity. Figure~\ref{fig:intro_head} illustrates the difference between existing credit assignment strategies and SHARP. The key technical innovation lies in our counterfactual masking mechanism, which mathematically isolates each agent’s causal impact by measuring the performance delta after removing relevant agents from the decision trajectory. 

SHARP ensures stable optimization by normalizing individual agent advantages across trajectory groups, yielding a low-variance gradient with coherent coordination. Across \emph{MuSiQue}, \emph{GAIA-text}, \emph{WebWalkerQA} and \emph{FRAMES} benchmarks, SHARP empirically achieves average match improvements of 23.66\% and 14.05\% over single-agent and multi-agent baselines, respectively. SHARP framework demonstrates strong cross-task generalization on \emph{DocMath-Eval} dataset and exhibits promising scaling laws, achieving 14.41 points of improvement over single-agent on an 8B backbone. Furthermore, coordination analysis reveals that SHARP reshapes the interaction structure by reducing the proportion of harmful subagents from 5.48\% to 4.40\%.

Our main contributions are summarized as follows:
\begin{itemize}

\item Proposing SHARP, a novel reinforcement learning framework to stabilize multi-agent training through precise credit attribution. 
By normalizing agent-specific advantages across trajectory groups, SHARP effectively facilitates stable policy gradients and provides unified solutions for aligning heterogeneous agents.

\item Introducing a flexible, tripartite reward decomposition mechanism comprising global task-alignment, Shapley-based marginal credit, and process-level execution validity signals.
This design applies to LLMs with varying sizes and diverse multi-agent structures, including sequential chains, communication graphs, and hierarchical planner-worker paradigms.

\item Conducting extensive experiments across diverse real-world benchmarks, demonstrating that SHARP with Qwen3-8B backbone achieves superior performance over state-of-the-art baselines, yielding an average match gains of 23.66\% and 14.05\% over single-agent and multi-agent approaches, respectively.

\end{itemize}

\section{Related Works}

\textbf{Multi-Agent Architectures and Operational Strategies.}
The transition from single-agent to multi-agent frameworks addresses critical limitations in tool-integrated planning and reasoning. While early methods such as ReAct \cite{yao2022react} use iterative prompting within a single agent, they often struggle with the cumulative noise introduced by complex, long-horizon tasks \cite{zhouwebarena}. To mitigate these challenges, recent research has introduced specialized collaborative structures that optimize information flow and task execution. For instance, COA \cite{li2025chain} employs a sequential strategy involving a chain of interacting agents to manage long-term context. Moreover, G-Designer \cite{zhang2025cu} and CARD \cite{anonymous2025card} utilize graph neural networks to approximate communication typologies, enabling efficient information exchange among agents. Recently, vertical architectures employing a \emph{planner-worker} or \emph{decomposer-solver} paradigm have become essential. For instance, AceSearcher \cite{xu2025acesearcher} and MATPO \cite{mo2025multi}, enhance multi-hop retrieval by training a unified LLM to alternate between decomposition and solving roles. 
Despite these architectural innovations, current research remains primarily focused on inference-time interaction protocols and topological stability, leaving the fundamental challenge of precise credit assignment during training largely unresolved.

\textbf{Reinforcement Learning and Credit Assignment.}
Reinforcement Learning (RL) is fundamental to aligning LLM agents with intricate tasks. Existing value-free approaches like GRPO \cite{shao2024deepseekmath} and One-Step Policy Optimization (OSPO) \cite{zhao2025one} are designed to mitigate critic estimation errors in highly uncertain environments. Building upon these foundations, contemporary extensions such as MATPO \cite{mo2025multi} and multi-agent GRPO \cite{hong2025multi} attempt to generalize group-relative baselines and hierarchical coordination to multi-agent settings. While these methods improve training stability, they often operate at the trajectory level and struggle to disentangle the specific marginal contributions of individual agents within a joint action space. 
Our work bridges this gap by incorporating counterfactual reasoning via Shapley-induced rewards, providing a mathematically grounded mechanism to isolate individual contributions within the optimization pipeline.

\section{Problem Setup}
\subsection{Tool-Integrated Interaction Trajectory}

We formalize tool-integrated reasoning as a sequential decision-making process where a language model interacts with external tools across multiple turns \cite{motwani2024malt,dong2025agentic}. Given an input query $q$, the model iteratively generates a sequence of actions and observes corresponding tool responses until reaching termination.

A tool-integrated trajectory $\tau$~\cite{qian2025toolrl} is defined as
\begin{equation}
\tau \triangleq \{a_1, s_1, a_2, s_2, \ldots, a_T, s_T\},
\end{equation}
where $a_t$ denotes a model action (e.g., reasoning or tool invocation) and $s_t$ is the corresponding tool observation.

At each step $t$, the action is sampled from a stochastic policy $\pi_\theta$ conditioned on the cumulative context as
$a_t \sim \pi_\theta(\cdot \mid p_{\mathrm{sys}}, q, a_{<t}, s_{<t})$
and receives a tool response
$s_t \sim P_{\mathrm{Tool}}(\cdot \mid a_t)$,
where $p_{\mathrm{sys}}$ specifies the system prompt and available tool schemas.

The joint trajectory probability factorizes as
\begin{equation}
P_\theta(\tau)=\prod_{t=1}^{T}\pi_\theta(a_t \mid p_{\mathrm{sys}}, q, a_{<t}, s_{<t})\, P_{\mathrm{Tool}}(s_t \mid a_t).
\end{equation}
Since tool responses are generated by an external process independent of $\theta$, optimization is driven solely by the generated actions $\{a_t\}_{t=1}^T$.

\subsection{Self-Play Multi-Agent Modeling}

We adopt a self-play paradigm via contextual conditioning~\cite{yang2025spell}, where a single shared policy $\pi_\theta$ instantiates heterogeneous roles (e.g., planner and worker) through role-specific system prompts, enabling parameter sharing across planning and execution.

The planner follows the conditional policy $\pi_\theta(\cdot \mid p_{\mathrm{planner}}, \cdot)$ and generates high-level actions as
\begin{equation}
a_t \sim \pi_\theta(\cdot \mid p_{\mathrm{planner}}, q, a_{<t}, s_{<t}).
\end{equation}

Each worker is instantiated by $\pi_\theta(\cdot \mid p_{\mathrm{worker}}, \cdot)$ and executes assigned subtasks. For subtask query $q_{\mathrm{subtask}\text{-}t}$, actions are sampled as
\begin{equation}
a_j^{t} \sim \pi_\theta(\cdot \mid p_{\mathrm{worker}}, q_{\mathrm{subtask}\text{-}t}, a^{t}_{<j}, s^{t}_{<j}).
\end{equation}

This self-play formulation integrates planning and execution trajectories under a shared policy, allowing the global objective to be jointly optimized across all roles.

\section{Methodology}

\begin{figure*}[t]
\centering
\includegraphics[width=0.9\textwidth]{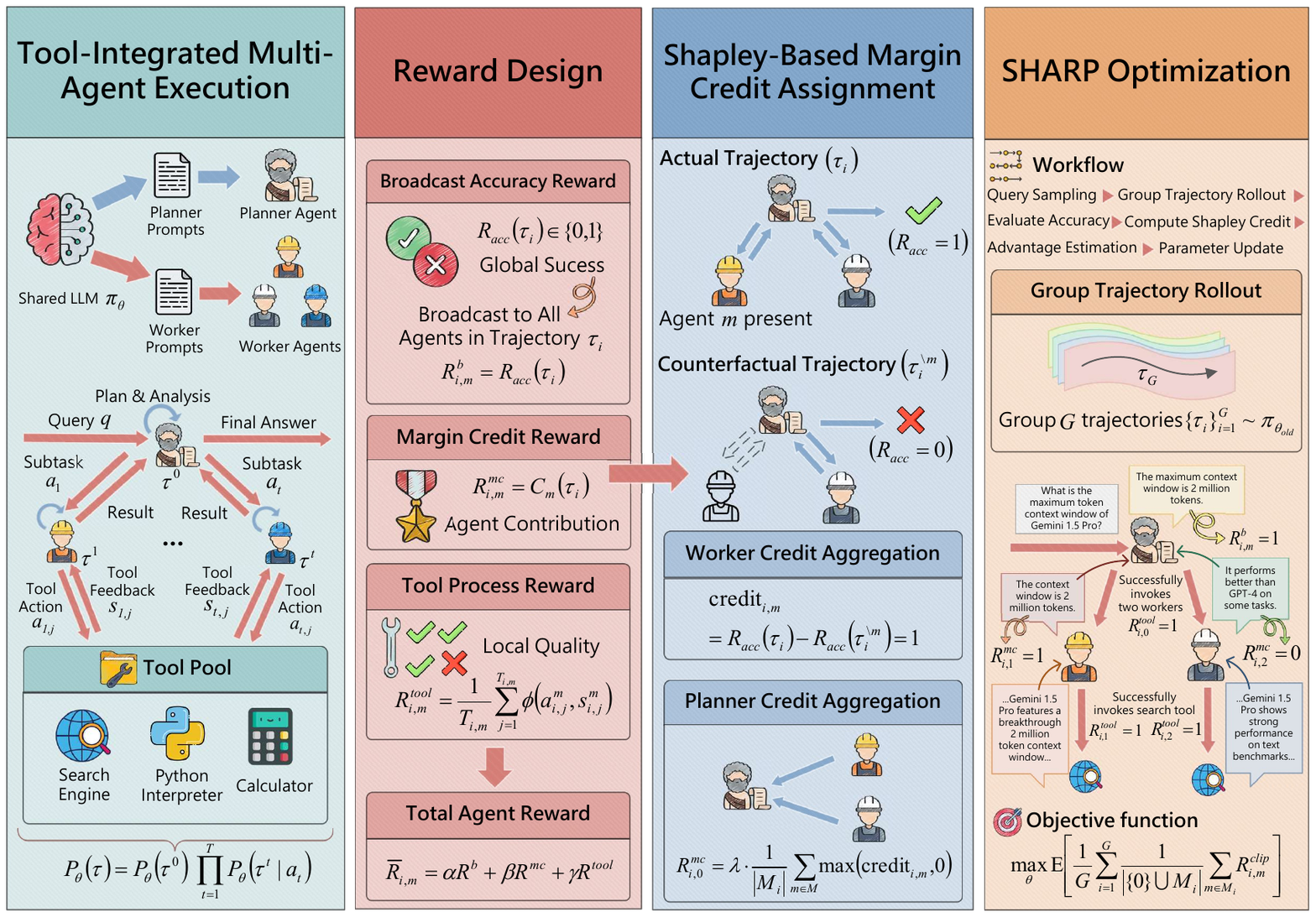}
\caption{Overview of SHARP framework. The pipeline involves (a) hierarchical interaction between planner and worker agents via a shared policy; (b) tripartite reward system integrating global accuracy, marginal credit, and tool process rewards; (c) marginal credit mechanism isolating agents' contribution via Shapley values; (d) SHARP workflow using group-relative policy for stable alignment.}
\label{fig:maingraph}
\end{figure*}

In this work, we study the Tool-Integrated Planning (TIP) paradigm, where a planner and a set of worker agents collaboratively solve user queries through multi-turn reasoning and tool use. Instead of maintaining separate policies, we adopt a \emph{parameter-sharing self-play} formulation, in which all roles are instantiated from a shared policy $\pi_\theta$ via role-specific prompts.
This design enables a unified model to capture both high-level planning and low-level execution behaviors. The overall workflow is summarized in Algorithm~\ref{sec:appendix-algorithm} and illustrated in Figure~\ref{fig:maingraph}.

\subsection{Tool-Integrated Multi-Agent Execution}

We model the execution process as a hierarchical decision-making procedure.
Given a query $q$, the shared policy $\pi_\theta$ produces a trajectory
$\tau = (\tau^0, \tau^1, \ldots, \tau^T)$,
where $\tau^0$ corresponds to the planner’s reasoning trace and each $\tau^t$ denotes the execution trajectory of the worker assigned to the $t$-th subtask.

Conditioned on the planner prompt $p_{\mathrm{planner}}$, the planner autoregressively generates actions $\{a_t\}_{t=1}^T$, each defining a subtask query $q_{\mathrm{subtask}\text{-}t}$ that triggers a worker execution loop.
Workers, prompted accordingly, interact with tools to complete the assigned subtasks.

The joint trajectory probability factorizes as
\begin{equation}
P_\theta(\tau)=P_\theta(\tau^0)\prod_{t=1}^{T}P_\theta(\tau^{t} \mid a_t),
\end{equation}
allowing the entire multi-agent process to be optimized end-to-end as a single trajectory.

\subsection{Compositional Reward Design}

Designing reward signals for collaborative MAS requires reconciling global task success with accurate agent-level feedback.
Broadcast rewards fail to disentangle individual contributions, while execution failures in tool-integrated settings cannot be addressed by terminal signals alone.
These challenges motivate a principled reward decomposition that ensures outcome alignment, faithful credit attribution, and execution validity.
Specifically, we introduce three axiomatic principles for precise reward assignment.

\begin{tcolorbox}[title=Multi-Agent Reward Assignment Rule]

In multi-agent reinforcement learning, the surrogate reward assigned to each participant $m\in\mathcal{M}$ should satisfy the following three principles:

\textbf{(i) Global Alignment (Effectiveness).}
The signal should be grounded in the final task outcome to align local decisions with overall trajectory success.

\textbf{(ii) Marginal Attribution (Credit Faithfulness).}
The signal should identify each agent's specific contribution, enabling precise credit assignment.

\textbf{(iii) Process Reward (Execution Validity).}
The signal should provide process-level feedback to ensure valid tool interactions.

\end{tcolorbox}

Following the above principles, we decompose the reinforcement signal of a TIP trajectory into three components:
\emph{broadcast accuracy reward}, \emph{marginal credit reward}, and \emph{tool process reward},
respectively capturing terminal success, agent-level contributions, and execution quality.

\textbf{Globally Broadcast Accuracy Reward.}
For a complete trajectory $\tau$, we define a binary terminal signal
\begin{equation}
R_{\mathrm{acc}}(\tau) \triangleq r_{\mathrm{acc}}(\tau) \in \{0,1\},
\end{equation}
which reflects whether the planner's final response aligns with the ground truth \cite{rafailov2023direct}.

Under the GRPO framework, $G$ trajectories $\{\tau_i\}_{i=1}^{G}$ are sampled per query, and the terminal reward is broadcast to the planner and all participating workers by
\begin{equation}
R^{\mathrm{b}}_{i,m} \triangleq R_{\mathrm{acc}}(\tau_i),
\quad m \in \{0\} \cup \mathcal{M}_i,
\end{equation}
where $\mathcal{M}_i$ denotes the set of workers involved in $\tau_i$.

\textbf{Marginal Credit Reward.}
Broadcast rewards alone cannot disentangle the contributions of different agents.
We therefore introduce a marginal credit reward that assigns agent-specific feedback based on their contribution to the final outcome:
\begin{equation}
R^{\mathrm{mc}}_{i,m} \triangleq \mathcal{C}_m\!\left(\tau_i;\, R_{\mathrm{acc}}\right),
\quad m \in \{0\} \cup \mathcal{M}_i,
\end{equation}
where $\mathcal{C}_m(\cdot)$ maps a multi-agent trajectory to per-agent credit conditioned on task success \cite{hong2024orpo}.
Concretely, $\mathcal{C}_m$ is instantiated via a counterfactual marginal contribution formulation, which quantifies how the task outcome changes when agent $m$ is ablated.
The exact definition is provided in Eq.~\eqref{eq:marginal-credit}.

\textbf{Tool Process Reward.}
For trajectory $\tau_i$, let $T_{i,m}$ be the number of tool calls raised by agent $m$. 
To ensure execution validity, we define the following process-level reward that evaluates the correctness of tool invocations:
\begin{equation}
R^{\text{tool}}_{i,m}
\triangleq
\begin{cases}
\displaystyle
\frac{1}{T_{i,m}}
\sum_{j=1}^{T_{i,m}} 
\phi\!\left(a^{m}_{i,j}, s^{m}_{i,j}\right),
& T_{i,m} > 0, \\[6pt]
0,
& T_{i,m} = 0,
\end{cases}
\end{equation}
\noindent
where $m \in \{0\} \cup \mathcal{M}_i$, and $\phi(\cdot)$ is a scalar function that evaluates the validity and executability of each tool usage.

\textbf{Aggregate Rewards.}
The total surrogate reward for optimizing agent $m$ on trajectory $\tau_i$ is a weighted combination of the aforementioned signals
\begin{equation*}
\bar{R}_{i,m} \triangleq \alpha \, R^{\mathrm{b}}_{i,m}
+ \beta \, R^{\mathrm{mc}}_{i,m} + \gamma \, R^{\mathrm{tool}}_{i,m}, 
\quad m \in \{0\} \cup \mathcal{M}_i,
\end{equation*}
where $\alpha$, $\beta$, and $\gamma$ calibrate the trade-offs between terminal effectiveness, credit faithfulness, and operational quality.

\subsection{Shapley-based Marginal Credit Assignment}

In this work, the credit functional is instantiated using an approximation of the Shapley value from cooperative game theory \cite{chen2023algorithms,napolitano2025lightningshap}. Given a set of agents $\mathcal{N}$ and a value function $v(\cdot)$, the Shapley value w.r.t. agent $m$ is formated by
\begin{equation*}
\begin{aligned}
\phi_m(v) &= \sum_{S \subseteq \mathcal{N} \setminus \{m\}}
\omega(S)\,( v(S \cup \{m\}) - v(S) ),\\
\text{where} \quad \omega(S) &= \frac{|S|!\, (|\mathcal{N}| - |S| - 1)!}{|\mathcal{N}|!}.
\end{aligned}
\end{equation*}
Notably, $\phi_m(v)$ quantifies the marginal contribution of agent $m \in \mathcal{N}$ across all possible coalitions.

Given the inherent complexity of multi-agent TIP, estimating the credits of all agents is computationally intractable. Therefore, we define the value function of a trajectory as
\begin{equation}
v(\tau) \triangleq R_{\mathrm{acc}}(\tau),
\end{equation}
which enables approximating individual Shapley values through a counterfactual marginal contribution framework.

Given a realized trajectory $\tau_i$ and a specific worker agent $m \in \mathcal{M}_i$, we quantify its credit by
\begin{equation}
\mathrm{credit}_{i,m} 
\triangleq R_{\mathrm{acc}}(\tau_i) - R_{\mathrm{acc}}(\tau_i^{\setminus m}),
\label{eq:marginal-credit}
\end{equation}
where $\tau_i^{\setminus m}$ represents the trajectory via ablating agent $m$ with masks and preserving the environment, other agents and their interactions. Consequently, the marginal credit reward for worker agents is assigned as
\begin{equation}
R^{\mathrm{mc}}_{i,m} \triangleq \mathrm{credit}_{i,m}.
\end{equation}

Unlike execution workers, whose actions directly affect the environment, the planner exerts a structural influence on terminal success through task decomposition. To formalize indirect causality, we define the planner's marginal credit as 
\begin{equation}
R^{\mathrm{mc}}_{i,0}\triangleq
\lambda \cdot \frac{1}{|\mathcal{M}_i|} \sum_{m \in \mathcal{M}_i} \max\!\left(\mathrm{credit}_{i,m},\, 0\right),
\end{equation}
where the mean operator ensures alignment of the planner's objective with the collective utility of its worker pool.

\begin{table*}[!t]
\centering
\small
\caption{
Comparison across benchmarks with respect to multi-agent support (MAS), training-based optimization (TRN), broadcast-only reward (BOR), and marginal credit modeling (MCR).
\cmark\ and \xmark\ indicate support and non-support.
$^{\dagger}$, $^{\ddagger}$, and no superscript denote methods using LLaMA-3.1-8B, Qwen2.5-7B, and Qwen3-8B, respectively.
We evaluate all methods on MuSiQue, GAIA-text, WebWalkerQA, and FRAMES, and report AVG as the average performance.
}
\setlength{\tabcolsep}{6pt}
\begin{tabular}{l|cccc|cccc|c}
\toprule
\rowcolor{headergray}
\textbf{Method} & \textbf{MAS} & \textbf{TRN}& \textbf{BOR} & \textbf{MCR}
& \textbf{MuSiQue} & \textbf{GAIA-text}
& \textbf{WebWalkerQA} & \textbf{FRAMES} & \textbf{AVG} \\
\midrule
\rowcolor{rowgray}
LLaMA-3.1-8B RAG                        & \xmark & \xmark & \xmark & \xmark & 7.20  & 8.82  & 0.77 & 5.81  & 5.65 \\
\rowcolor{rowgray}
Qwen3-8B RAG                       & \xmark & \xmark & \xmark & \xmark & 8.60  & 15.40 & 1.23 & 6.78  & 8.00 \\
\midrule
Plan-Search $^{\dagger}$          & \xmark & \xmark & \xmark & \xmark & 26.66 & 10.04 & 3.32 & 10.76 & 12.70 \\
Plan-Search       & \xmark & \xmark & \xmark & \xmark & 36.35 & 27.48 & 6.77 & 28.48 & 24.77 \\
Search-R1 $^{\ddagger}$                    & \xmark & \cmark &  \xmark & \xmark & 18.11 & 14.15 & 2.30 & 11.30 & 11.47 \\
Single-agent GRPO             & \xmark & \cmark &  \xmark & \xmark & 45.93 & 27.97 & 7.47 & 30.20 & 27.89 \\
\midrule
\rowcolor{rowgray}
Hierarchical MAS $^{\dagger}$     & \cmark & \xmark &  \xmark & \xmark & 35.22 & 13.36 & 5.57 & 21.21 & 18.84 \\
\rowcolor{rowgray}
Hierarchical MAS     & \cmark & \xmark &  \xmark & \xmark & 38.23 & 27.53 & 7.42 & 32.18 & 26.34 \\
\midrule
G-Designer                   & \cmark & \cmark & \xmark & \xmark & 38.50 & 28.15 & 4.70 & 28.28 & 24.90 \\
CARD                          & \cmark & \cmark &  \xmark &\xmark & 45.00 & 32.89 & 7.38 & 27.31 & 28.15 \\
COA                           & \cmark & \cmark &  \xmark &\xmark & 44.28 & 32.00 & 7.22 & 32.10 & 28.90 \\
AceSearcher$^{\dagger}$                  & \cmark & \cmark & \cmark & \xmark & 36.41 & 20.05 & 7.04 & 27.38 & 22.72 \\
MATPO                         & \cmark & \cmark & \cmark & \xmark & 47.00 & 31.65 & 7.47 & 37.10 & 30.81 \\
\midrule
\rowcolor{rowgray}
\textbf{SHARP $^{\dagger}$}      & \cmark & \cmark & \cmark &\cmark & 46.14 & 23.23 & 7.60 & 25.71 & 25.67 \\
\rowcolor{rowgray}
\textbf{SHARP }         & \cmark & \cmark & \cmark &\cmark & \textbf{50.76} & \textbf{33.70} & \textbf{8.50} & \textbf{37.29} & \textbf{32.56} \\
\bottomrule
\end{tabular}
\label{tab:main_results}
\vspace{-0.4cm}
\end{table*}

\subsection{SHARP Optimization via GRPO}
To achieve efficient, stable optimization in the presence of sparse rewards and multi-agent coordination, we use the improved group relative policy gradient \cite{nauman2023many,shao2024deepseekmath} to normalize agentic rewards. 

\textbf{Agentic Group Relative Advantage.}
For $G$ sampled trajectories $\{\tau_i\}_{i=1}^{G}$, we collect the corresponding rewards $\{\bar{R}_{i,m}\}_{i=1}^{G}$, and compute the mean $\mu_m$ and standard deviation $\sigma_m$ for each individual agent $m$.
The group relative advantage $\hat{A}_{i,m}$ of agent $m$ on trajectory $\tau_i$ is defined as
\begin{equation}
\hat{A}_{i,m}
\triangleq
\frac{\bar{R}_{i,m} - \mu_m}{\sigma_m + \delta},
\end{equation}
where $\delta$ is a constant controlling the stability.

\textbf{SHARP Clipped Surrogate Objective.}
Let $\{a^{m}_{i,j}\}_{j=1}^{T_{i,m}}$ be the sequence of actions, and let $\mathrm{ctx}^{m}_{i,j}$ be the corresponding context generated by agent $m$ on trajectory $\tau_i$. 
We define the ratio between the current and old policies as
\begin{equation}
\begin{aligned}
R_{i,m}\triangleq \exp\Bigg(
&\sum_{j=1}^{T_{i,m}} \log \pi_{\theta}\!\left(a^{m}_{i,j}\mid \mathrm{ctx}^{m}_{i,j}\right)\\
&- \sum_{j=1}^{T_{i,m}} \log \pi_{\theta_{\mathrm{old}}}\!\left(a^{m}_{i,j}\mid \mathrm{ctx}^{m}_{i,j}\right) \Bigg).
\end{aligned}
\end{equation}

The clipped surrogate objective w.r.t. agent $m$ and trajectory $\tau_i$ is formulated by
\begin{equation*}
R^{\mathrm{clip}}_{i,m}\triangleq
\min\!\left(R_{i,m}\,\hat{A}_{i,m},\mathrm{clip}(R_{i,m},\,1-\epsilon,\,1+\epsilon)\,\hat{A}_{i,m}\right).
\end{equation*}

\textbf{Overall Optimization Objective.}
The SHARP objective $J_{\mathrm{SHARP}}(\theta)$ aggregates clipped advantages across trajectories and agents within the group $\{0\}\cup\mathcal{M}_i$, where
\begin{equation*}
J_{\mathrm{SHARP}}(\theta) 
\triangleq \mathbb{E}\!\left[ \frac{1}{G} \sum_{i=1}^{G} \frac{1}{|\{0\}\cup\mathcal{M}_i|} \sum_{m\in\{0\}\cup\mathcal{M}_i} R^{\mathrm{clip}}_{i,m} \right],
\end{equation*}
where policy parameters $\theta$ are updated by maximizing $J_{\mathrm{SHARP}}(\theta)$, enabling the coordinated refinement of planning and execution capabilities. Corresponding SHARP optimization steps are present in Algorithm \ref{alg:sharp_tip}, Appendix \ref{sec:appendix-algorithm}.

\section{Experimental Evaluation}

\newlength{\frameheight}
\setlength{\frameheight}{0.16\textheight}

We conduct a series of experiments to evaluate the effectiveness, stability, scalability, and interpretability of the proposed SHARP framework. Our empirical investigation is structured around four primary research questions:
\textbf{RQ1:} How does SHARP perform compared to existing single-agent and multi-agent baselines across diverse benchmarks?
\textbf{RQ2:} Which components of Shapley-based credit assignment contribute most to the performance gains?
\textbf{RQ3:} How stable and scalable is SHARP across heterogeneous tasks, model sizes, and training steps?
\textbf{RQ4:} How does SHARP characterize and reshape planner-worker coordination?
For extended analyses, please refer to Appendices \ref{sec:cost-analysis}, \ref{sec:forget-analysis}, and \ref{sec:appendix-case}.

\subsection{Settings}
\textbf{Benchmarks and Metrics.}
We utilize five distinct benchmarks, including MuSiQue \cite{trivedi2022musique}, GAIA-text \cite{mialon2023gaia}, WebWalkerQA \cite{wu2025webwalker}, FRAMES \cite{krishna2025fact} and DocMath-Eval \cite{zhao2024docmath}. Please see Appendix~\ref{sec:appendix-data} for more details.

\textbf{Baselines and Setup.}
We compare our method against four categories of baselines.
(i) \textbf{Vanilla LLMs}, including Llama-3.1-it 8B \cite{grattafiori2024llama} and Qwen-3 8B \cite{yang2025qwen3}.
(ii) \textbf{Prompt based planning and search methods}, such as Plan-Search \cite{yu2023prompt}.
(iii) \textbf{Single agent reinforcement learning methods}, including Search R1 \cite{jin2025search} and Single agent GRPO \cite{shao2024deepseekmath}.
(iv) \textbf{Multi-agent systems without marginal credit modeling}, including \emph{planner-worker} frameworks as well as dynamic graph-based frameworks such as G-designer \cite{zhang2025cu} and CARD \cite{anonymous2025card}, and reinforcement trained methods such as AceSearcher \cite{xu2025acesearcher}, COA \cite{li2025chain}, and MATPO \cite{mo2025multi}.
Please see Appendix~\ref{sec:appendix-impl} for more details.

\subsection{RQ1: Overall Performance}

\textbf{Motivation:} Multi-agent LLM systems vary across intertwined factors such as architecture, training-based optimization, and reward design, making the source of performance gains unclear. A controlled comparison is therefore required to assess the contribution of marginal credit modeling.

\textbf{Observation 1: Marginal credit modeling consistently delivers the best overall performance.}
As shown in Table~\ref{tab:main_results}, SHARP (Qwen) achieves the highest average performance of 32.56, surpassing the strongest non-marginal multi-agent baseline MATPO by 1.75 points. 
Overall, SHARP yields average match gains of 23.66\% and 14.05\% over single-agent and multi-agent baselines, respectively(see Appendix~\ref{sec:appendix-impl} for detailed computation), demonstrating consistent improvements across diverse benchmarks.
Notably, SHARP outperforms structured but untrained Planner--Worker methods by over 6.22 points, and ranks first on all evaluated tasks, including a score of 50.76 on MuSiQue.
Taken together, these results indicate that explicitly modeling marginal credit is a key factor behind SHARP’s consistent gains, beyond architectural design or optimization strategy alone.

\begin{figure*}[!t]
\centering

\begin{tikzpicture}[baseline=(L.base)]
\node[
  draw=gray,
  rounded corners=6pt,
  inner sep=6pt,
  line width=0.8pt
] (L) {
  \begin{minipage}[c][\frameheight][c]{0.62\linewidth}
    \centering
    \vspace{10pt}
    \includegraphics[width=0.48\linewidth]{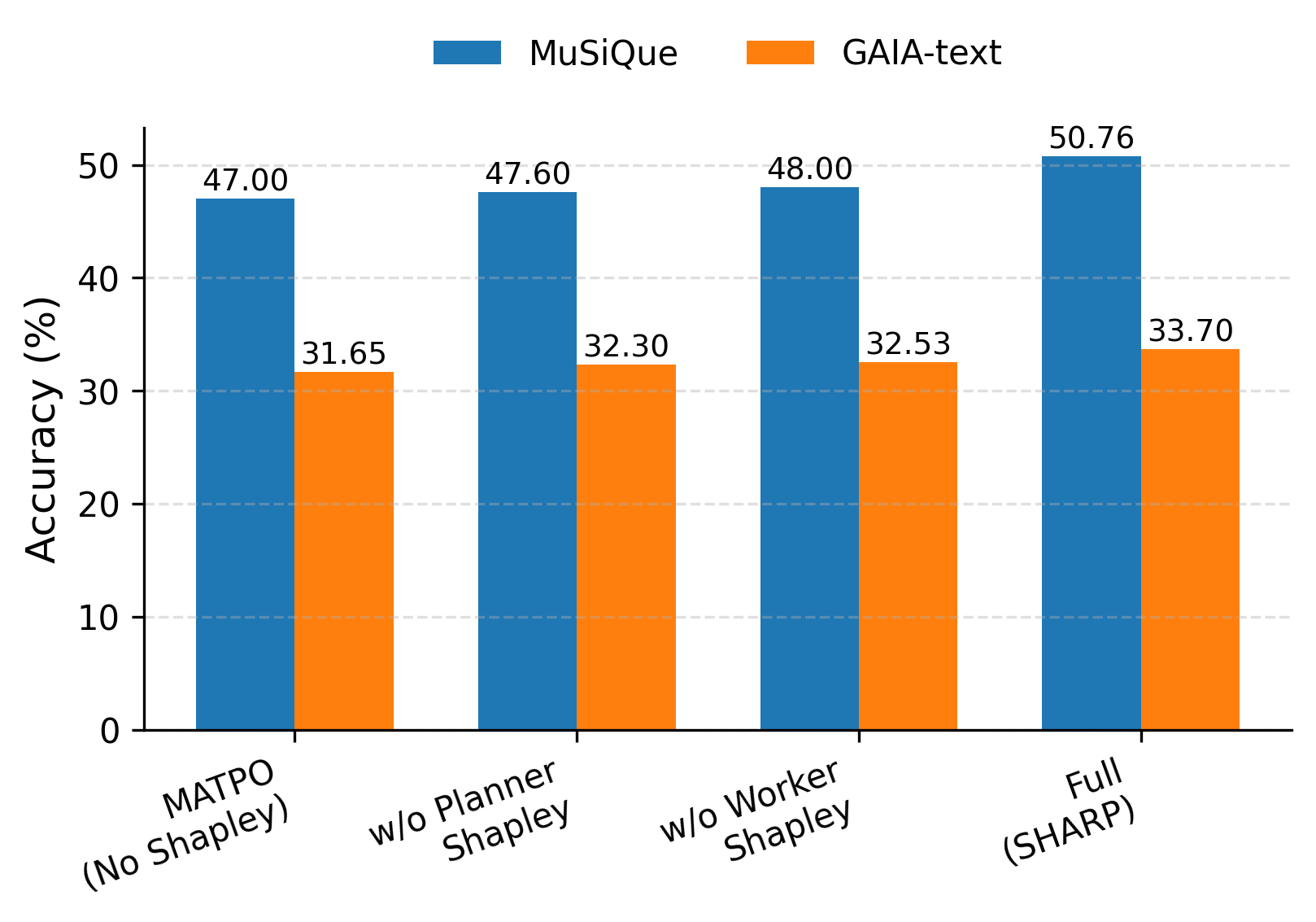}
    \hfill
    \includegraphics[width=0.48\linewidth]{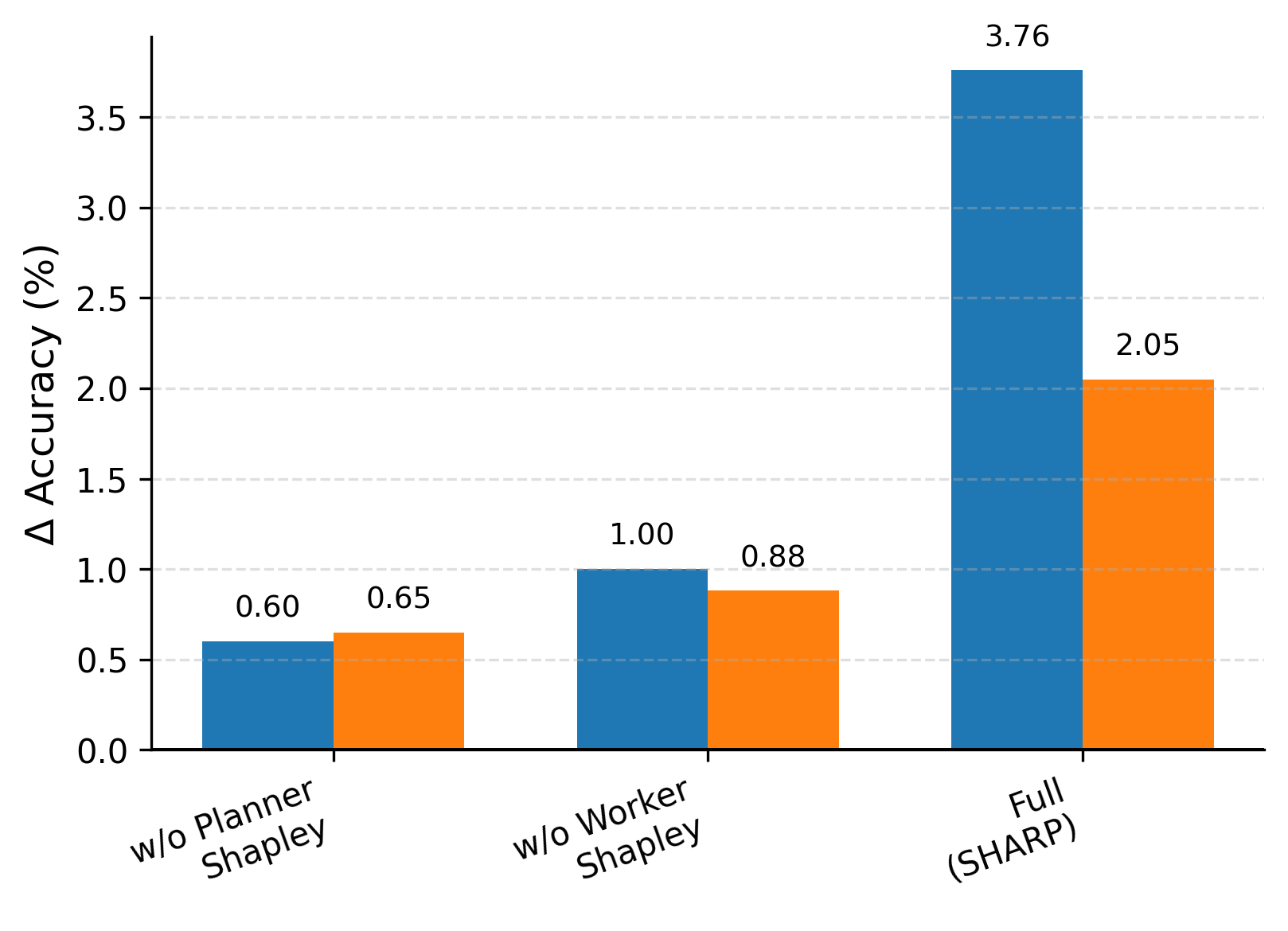}

    \vspace{2pt}

  \end{minipage}
};
\end{tikzpicture}
\hfill
\begin{tikzpicture}[baseline=(R.base)]
\node[
  draw=gray,
  rounded corners=6pt,
  inner sep=6pt,
  line width=0.8pt
] (R) {
  \begin{minipage}[c][\frameheight][c]{0.32\linewidth}
    \centering
    \includegraphics[height=1\frameheight]{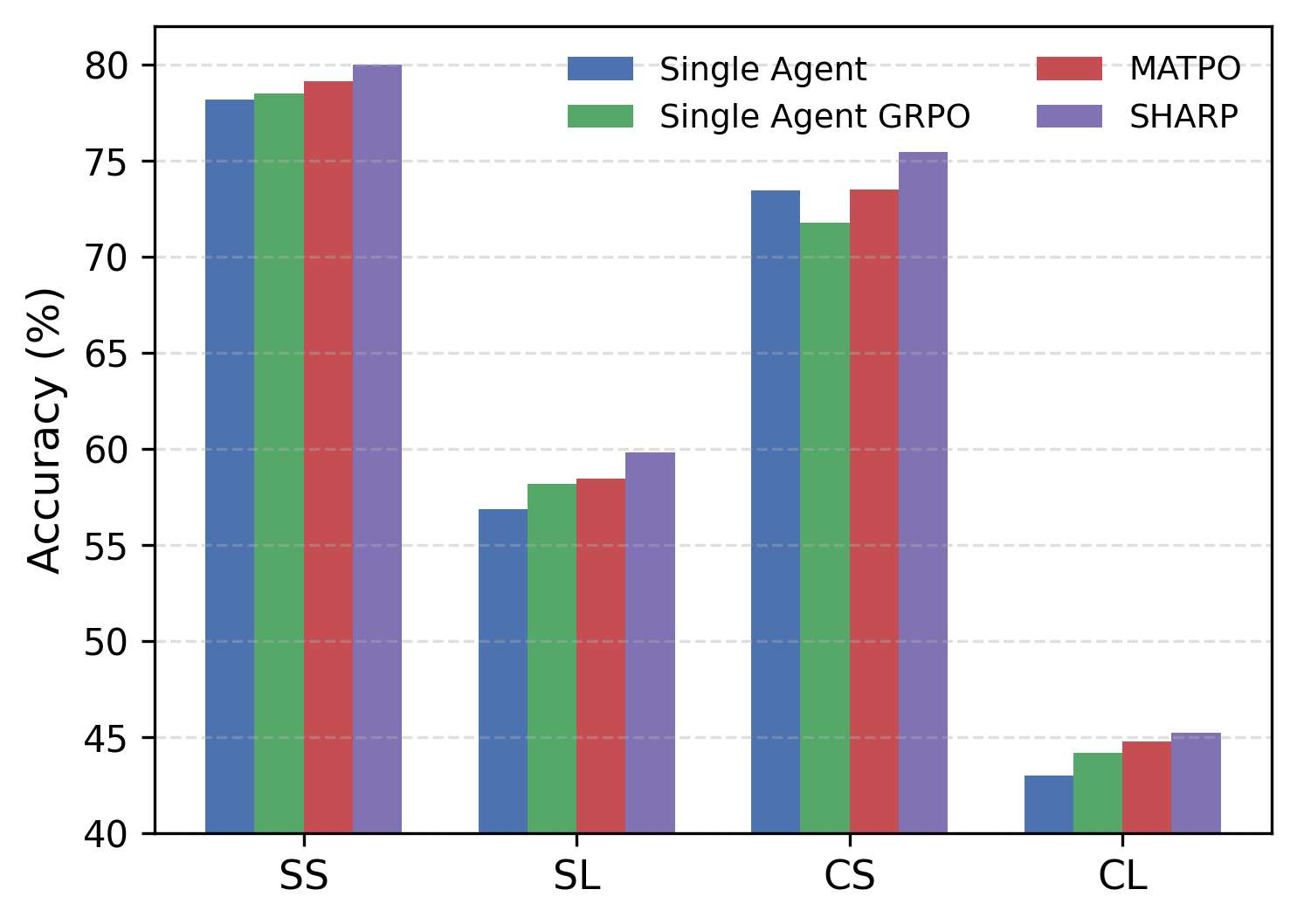}

    \vspace{2pt}

  \end{minipage}
};
\end{tikzpicture}

\caption{\textbf{Left}: Ablation studies on MuSiQue and GAIA-text comparing full SHARP with variants that remove planner-level or worker-level Shapley credit.
\textbf{Middle}: The corresponding accuracy differences ($\Delta$ Accuracy) measured relative to the no-Shapley baseline on each benchmark.
\textbf{Right}: Evaluation on DocMath-Eval across four document-level reasoning settings, including Simple-Short (SS), Simple-Long (SL), Complex-Short (CS), and Complex-Long (CL).}
\label{fig:component_generalization}
\end{figure*}

\textbf{Observation 2: Optimization strategy outperforms structural complexity alone.} The results in Table \ref{tab:main_results} further reveal a progressive performance trend: Zero-shot RAG baselines perform poorly (5.65 - 8.00), while an untrained structured design (e.g., Plan-Search) improves scores to 12.70 - 24.77.
Introducing reinforcement learning at the single-agent level (Single-agent GRPO) further boosts the average score to 27.89.
Multi-agent execution without training (Planner--Worker) yields limited gains (18.84--26.34), while MAS RL baselines lacking marginal credit modeling (e.g., CARD, COA, MATPO) consistently reach \textbf{28.15--30.81}.
This also indicates that both MAS architectures and reinforcement learning are crucial for enhancing planning and execution capabilities in realistic scenarios.

\begin{insightbox}
    \textbf{Insight 1:} Marginal credit modeling provides a stronger performance boost than architectural structure or optimization strategy alone, consistently yielding superior overall accuracy across benchmarks in heterogeneous multi-agent LLM systems.
\end{insightbox}

\subsection{RQ2: Component Contribution Assessment}
\textbf{Motivation:} In hierarchical multi-agent systems, planner and worker agents play distinct yet interdependent roles, making their individual contributions to overall performance unclear.
Ablation analysis is therefore required to disentangle planner- and worker-level credit assignment relative to the full SHARP model and standard baselines.

\begin{figure}[]
\centering
\begin{tikzpicture}
\node[
  draw=gray,              
  line width=0.8pt,       
  rounded corners=4pt,   
  inner sep=6pt          
] {
  \includegraphics[width=0.85\linewidth]{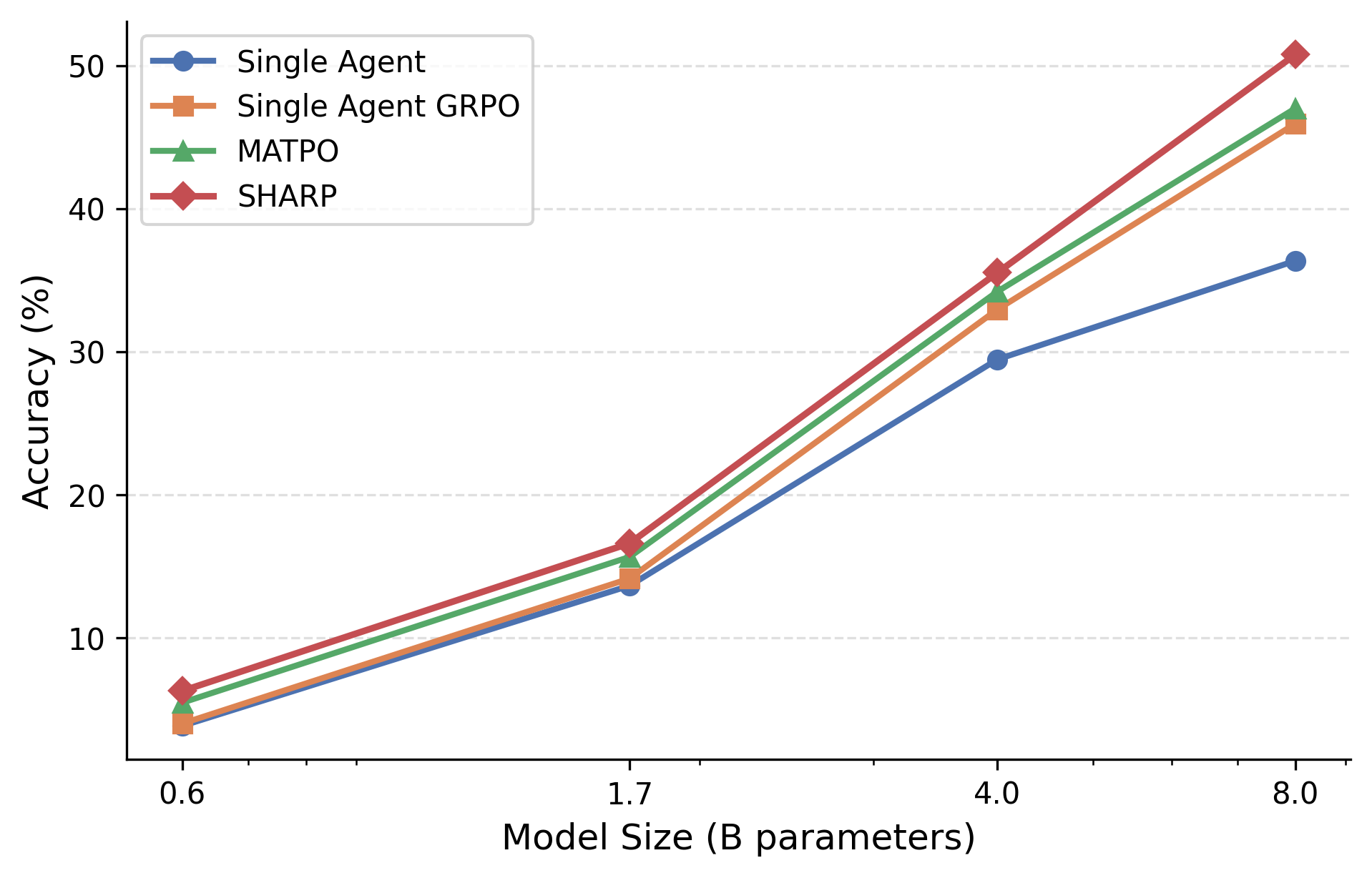}
};
\end{tikzpicture}

\caption{Parameter scalability on MuSiQue from 0.6B to 8B.
SHARP shows consistent improvement as the model size increases and achieves a larger advantage over the baselines at larger scales.}
\label{fig:musique_scaling}
\vspace{-0.5cm}
\end{figure}

\textbf{Observation 1: Mutual gains through joint training.}
As shown in Figure~\ref{fig:component_generalization} (left),
The results show a clear benefit when rewards are given to both roles simultaneously. For example, on MuSiQue, accuracy increases from a 47.00 baseline to 47.60 (48.00) with planner-only (worker-only) credit, but reaches 50.76 with the complete SHARP model. By aligning reward signals across all roles, SHARP improves high-level planning, tool execution, and internal interactions together, resulting in a final performance that is much higher than the sum of separate updates.

\textbf{Observation 2: Planning strategy and execution quality.}
As illustrated in Figure~\ref{fig:component_generalization} (middle),
Planner's credit primarily refines task decomposition logics, yielding empirical gains of 0.6 to 0.65. 
Moreover, workers' credits enable the identification of high-utility and redundant tool calls, yielding larger boosts of 0.88 to 1.00. This indicates that, in long-horizon scenarios, tool-calling and execution failures are sometimes more challenging for multi-agent training. 

\begin{insightbox}
    \textbf{Insight 2:} Coordinated credit assignment across planner and worker agents yields synergistic performance gains, where planner-level credit improves task decomposition while worker-level credit more effectively mitigates execution-level failures in long-horizon settings.
\end{insightbox}

\subsection{RQ3: Stability and Scalability Analysis}

\textbf{Motivation:} Practical MAS require robustness to task heterogeneity, model scaling and long-horizon training, motivating a systematic analysis of stability and scalability.

\textbf{Observation 1: Generalization across diverse reasoning tasks.}
To better evaluate SHARP's generalization across different task distributions, we conduct experiments on DocMath-Eval with four document-level reasoning tasks, including simple short (SS), simple long (SL), complex short (CS) and complex long (CL) scenarios \cite{zheng2025knowledge,tang2025finmmr}. As shown in Figure~\ref{fig:component_generalization} (right), SHARP achieves the best performance on all subtasks, verifying its strong generalization capability across various task distributions and heterogeneous reasoning settings.

\textbf{Observation 2: Scaling with model parameters.}
We examined how performance scales with model size by testing Qwen3 backbones ranging from 0.6B to 8B on MuSiQue. Across this spectrum, accuracy for the single-agent baseline rises from 3.85 to 36.35, while Single-agent GRPO and MATPO improve from 4.01 and 5.47 at 0.6B to 45.93 and 47.00 at 8B, respectively. Figure \ref{fig:musique_scaling} shows that SHARP scales more effectively than these baselines, where the performance grows from 6.29 at 0.6B to 50.76 at 8B. Notably, the difference between SHARP and single-agent widens with larger model sizes, reaching 14.41 points at the 8B scale. This trend verifies that SHARP becomes more effective on stronger backbones.

\begin{figure}[t]
\centering
\includegraphics[width=0.9\linewidth]{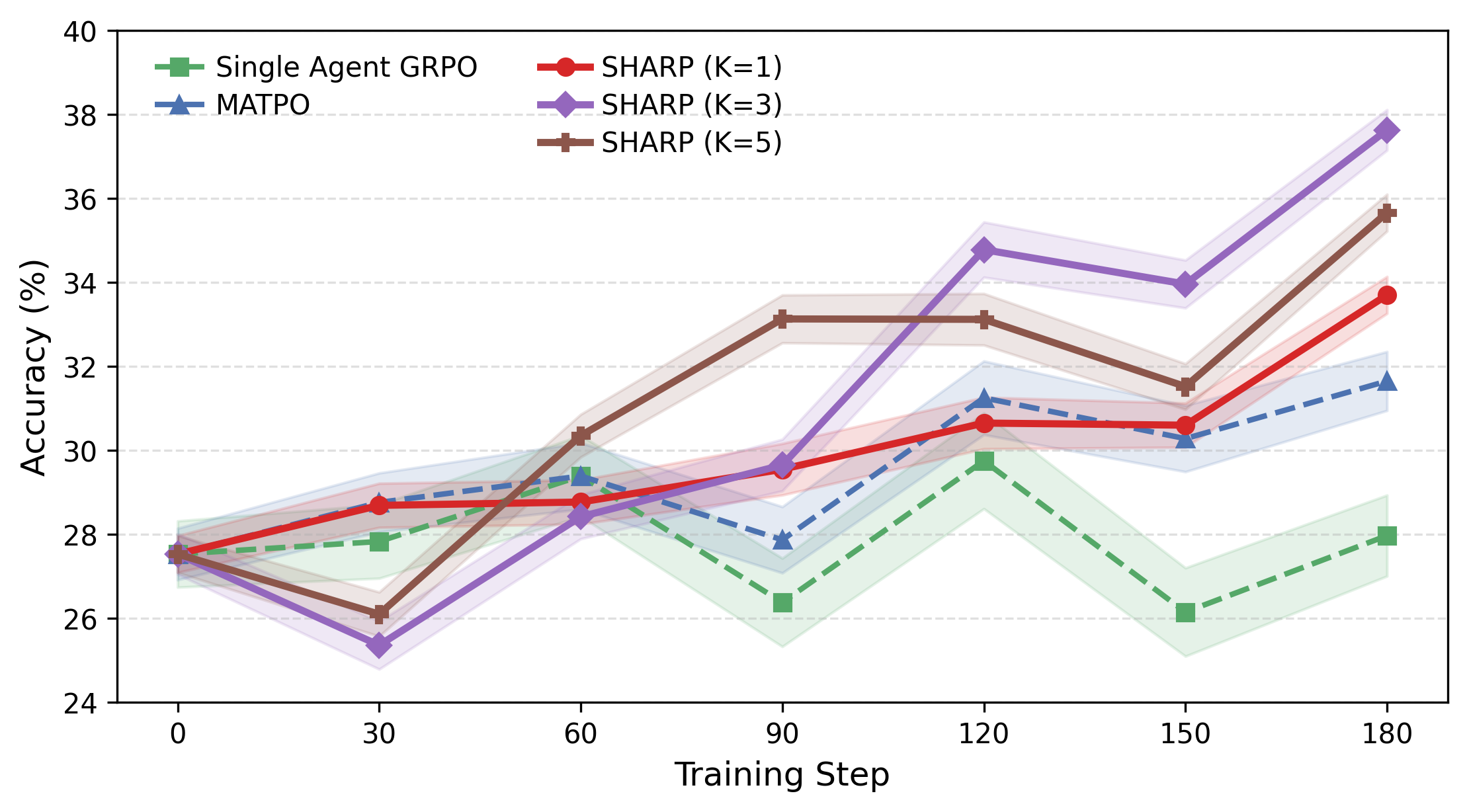}
\caption{Training-step scalability on GAIA-text from 0 to 180 steps.
SHARP improves steadily as training progresses and avoids the instability observed in the baseline; shaded areas denote 95\% confidence intervals.}
\label{fig:gaia_steps}
\vspace{-0.5cm}
\end{figure}

\textbf{Observation 3: Training stability and budget scaling.}
We analyzed the convergence behavior on GAIA-text over 180 optimization steps in Figure \ref{fig:gaia_steps}. While single-agent GRPO and MATPO exhibit fluctuations, SHARP shows a monotonic improvement in accuracy (from 27.53 to 33.70). This stable progress suggests that explicit credit modeling mitigates the reward variance frequently present in multi-agent RL, leading to more reliable training over longer horizons.

\begin{insightbox}
    \textbf{Insight 3:} Explicit marginal credit modeling yields robust generalization across task distributions, scales effectively with model size, and stabilizes long-horizon training, indicating that SHARP benefits more from stronger backbones and extended optimization.
\end{insightbox}

\subsection{RQ4: Coordination Analysis}

\textbf{Motivation:} Training strategies reshape coordination in multi-agent systems, but their impact on planner--worker interactions remains unclear. We analyze coordination using Shapley-based credit signals, defining the planner score as the planner’s average Shapley value and labeling subagents with positive (negative) credit as useful (harmful).

\begin{figure}[t]
\centering
\includegraphics[width=0.85\linewidth]{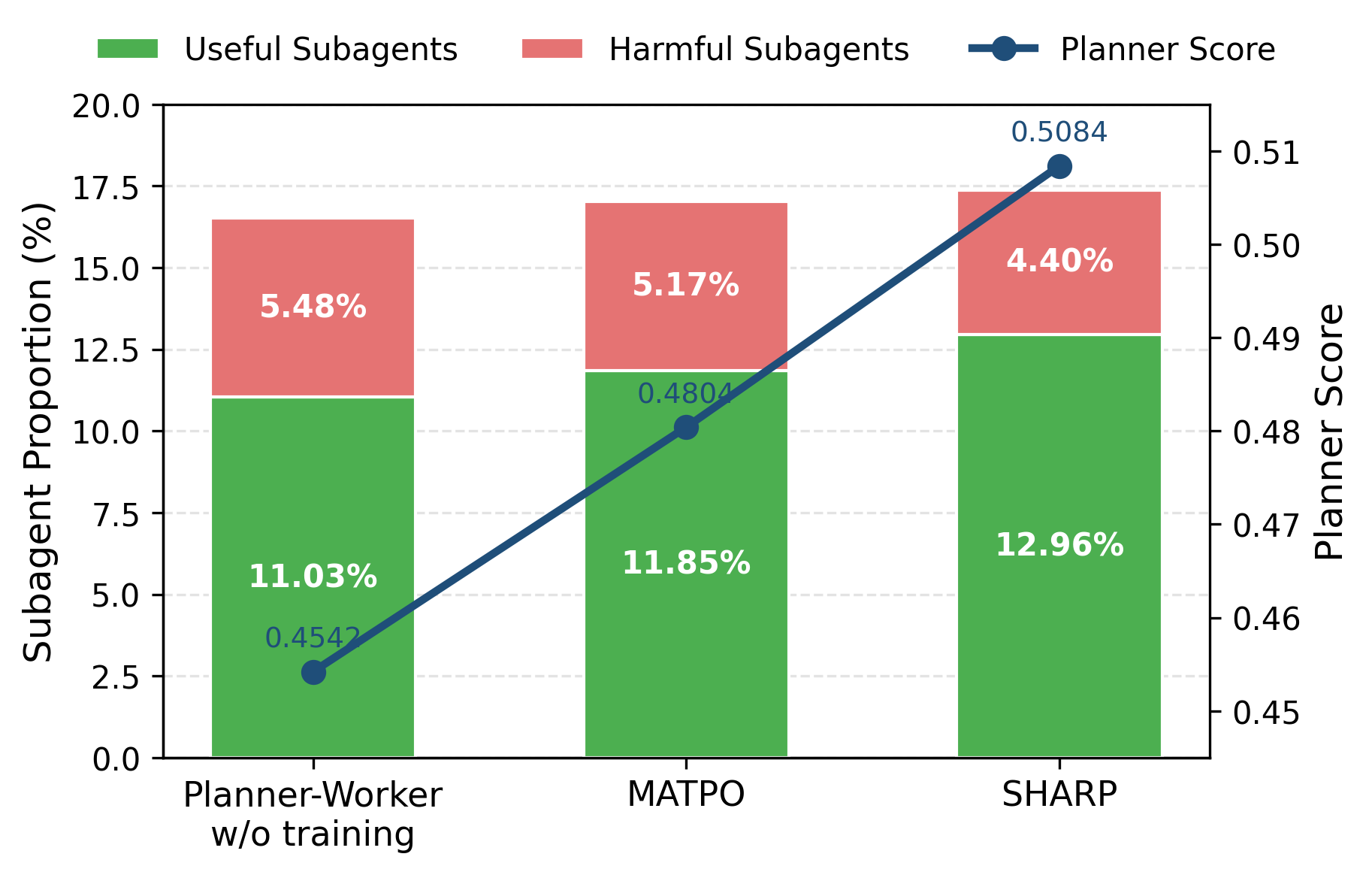}
\caption{Coordination metrics based on marginal Shapley credit.
The planner score is defined as the planner’s average Shapley value.
Subagents with positive marginal Shapley credit are categorized as useful, while those with negative credit are categorized as harmful.}
\label{fig:shapley_coordination}
\vspace{-0.5cm}
\end{figure}

\textbf{Observation 1: Improved planning and subagent selection.}
Figure~\ref{fig:shapley_coordination} shows the planner scores and subagent utility across different training schemes. SHARP increases the planner score to 0.5084, outperforming both the vanilla baseline (0.4542) and MATPO (0.4804). Simultaneously, the proportion of useful subagents increases from 11.03\% to 12.96\%. These results indicate that SHARP forces the planner to generate more effective subtasks and to select workers more strategically.

\textbf{Observation 2: Filtering harmful interactions.}
A key benefit of marginal credit assignment is its ability to identify and penalize counterproductive actions. SHARP reduced the proportion of harmful subagent calls \cite{chan2025infrastructure} from 5.48\% to 4.40\%. By assigning lower credit to workers who degrade the final outcome, the framework effectively cleans the coordination trace and encourages more reliable agent interactions. A qualitative case study illustrating useful and harmful subagent behaviors is in Appendix~\ref{sec:appendix-case}.

\textbf{Observation 3: Common limits in systemic efficiency.}
Despite these improvements, our analysis reveals a broader, common challenge: \emph{useful} subagents remain a minority of total calls in the tested systems. This confirms that a significant portion of current multi-agent tool usage could be redundant or neutral. We believe this observation illustrates a fundamental bottleneck in multi-agent designs, highlighting the need for future research into more aggressive pruning of low-utility execution paths.

\begin{insightbox}
    \textbf{Insight 4:} Shapley-based marginal credit modeling reshapes planner–worker coordination by improving subtask selection, filtering harmful interactions, and exposing inefficiencies in multi-agent coordination.
\end{insightbox}

\section{Conlucions}

In this paper, we introduce the Shapley-based Hierarchical Attribution for Reinforcement Policy (SHARP), a mathematically grounded framework designed to address the fundamental, precise credit-assignment bottleneck in multi-agent systems. By leveraging a tripartite decomposed reward mechanism, SHARP effectively disentangles the individual causal impact of agents from confounded aggregate outcomes, providing a precise learning signal for parameter-sharing MAS hierarchies. This work presents fine-grained credit attribution for stable joint optimization and cross-task generalization in complex multi-agent systems. More broadly, SHARP provides a principled foundation for scalable and interpretable coordination in hierarchical MAS.

\section*{Impact Statement}
This paper presents work whose goal is to advance the field of machine learning, particularly in multi-agent reinforcement learning with tool-integrated planning. The proposed SHARP framework integrates Shapley-based credit attribution with hierarchical planner–worker optimization to address the challenge of precise credit assignment in collaborative multi-agent systems. 
Following this direction, the framework is broadly applicable to other multi-agent learning settings that require stable joint optimization and interpretable agent coordination. There may be societal consequences of our work, none of which we feel need to be specifically highlighted here.
\bibliography{icml2026}

@inproceedings{
anonymous2025card,
title={{CARD}: Towards Conditional Design of Multi-agent Topological Structures},
author={Anonymous},
booktitle={The Fourteenth International Conference on Learning Representations},
year={2026},
url={https://openreview.net/forum?id=JgvJdICc6P}
}

@article{li2025chain,
  title={Chain-of-agents: End-to-end agent foundation models via multi-agent distillation and agentic rl},
  author={Li, Weizhen and Lin, Jianbo and Jiang, Zhuosong and Cao, Jingyi and Liu, Xinpeng and Zhang, Jiayu and Huang, Zhenqiang and Chen, Qianben and Sun, Weichen and Wang, Qiexiang and others},
  journal={arXiv preprint arXiv:2508.13167},
  year={2025}
}

@article{qian2025toolrl,
  title={Toolrl: Reward is all tool learning needs},
  author={Qian, Cheng and Acikgoz, Emre Can and He, Qi and Wang, Hongru and Chen, Xiusi and Hakkani-T{\"u}r, Dilek and Tur, Gokhan and Ji, Heng},
  journal={arXiv preprint arXiv:2504.13958},
  year={2025}
}

@article{lewis2020retrieval,
  title={Retrieval-augmented generation for knowledge-intensive nlp tasks},
  author={Lewis, Patrick and Perez, Ethan and Piktus, Aleksandra and Petroni, Fabio and Karpukhin, Vladimir and Goyal, Naman and K{\"u}ttler, Heinrich and Lewis, Mike and Yih, Wen-tau and Rockt{\"a}schel, Tim and others},
  journal={Advances in neural information processing systems},
  volume={33},
  pages={9459--9474},
  year={2020}
}

@inproceedings{yao2022react,
  title={React: Synergizing reasoning and acting in language models},
  author={Yao, Shunyu and Zhao, Jeffrey and Yu, Dian and Du, Nan and Shafran, Izhak and Narasimhan, Karthik R and Cao, Yuan},
  booktitle={The eleventh international conference on learning representations},
  year={2022}
}

@article{dong2025agentic,
  title={Agentic reinforced policy optimization},
  author={Dong, Guanting and Mao, Hangyu and Ma, Kai and Bao, Licheng and Chen, Yifei and Wang, Zhongyuan and Chen, Zhongxia and Du, Jiazhen and Wang, Huiyang and Zhang, Fuzheng and others},
  journal={arXiv preprint arXiv:2507.19849},
  year={2025}
}

@article{li2025webthinker,
  title={Webthinker: Empowering large reasoning models with deep research capability},
  author={Li, Xiaoxi and Jin, Jiajie and Dong, Guanting and Qian, Hongjin and Wu, Yongkang and Wen, Ji-Rong and Zhu, Yutao and Dou, Zhicheng},
  journal={arXiv preprint arXiv:2504.21776},
  year={2025}
}

@article{rafailov2023direct,
  title={Direct preference optimization: Your language model is secretly a reward model},
  author={Rafailov, Rafael and Sharma, Archit and Mitchell, Eric and Manning, Christopher D and Ermon, Stefano and Finn, Chelsea},
  journal={Advances in neural information processing systems},
  volume={36},
  pages={53728--53741},
  year={2023}
}

@article{hong2024orpo,
  title={Orpo: Monolithic preference optimization without reference model},
  author={Hong, Jiwoo and Lee, Noah and Thorne, James},
  journal={arXiv preprint arXiv:2403.07691},
  year={2024}
}

@article{schulman2017proximal,
  title={Proximal policy optimization algorithms},
  author={Schulman, John and Wolski, Filip and Dhariwal, Prafulla and Radford, Alec and Klimov, Oleg},
  journal={arXiv preprint arXiv:1707.06347},
  year={2017}
}

@article{hong2025multi,
  title={Multi-Agent Deep Research: Training Multi-Agent Systems with M-GRPO},
  author={Hong, Haoyang and Yin, Jiajun and Wang, Yuan and Liu, Jingnan and Chen, Zhe and Yu, Ailing and Li, Ji and Ye, Zhiling and Xiao, Hansong and Chen, Yefei and others},
  journal={arXiv preprint arXiv:2511.13288},
  year={2025}
}

@article{motwani2024malt,
  title={Malt: Improving reasoning with multi-agent llm training},
  author={Motwani, Sumeet Ramesh and Smith, Chandler and Das, Rocktim Jyoti and Rafailov, Rafael and Laptev, Ivan and Torr, Philip HS and Pizzati, Fabio and Clark, Ronald and de Witt, Christian Schroeder},
  journal={arXiv preprint arXiv:2412.01928},
  year={2024}
}

@article{zhao2025one,
  title={One step is enough: Multi-agent reinforcement learning based on one-step policy optimization for order dispatch on ride-sharing platforms},
  author={Zhao, Zijian and Li, Sen},
  journal={arXiv preprint arXiv:2507.15351},
  year={2025}
}

@inproceedings{zhouwebarena,
  title={WebArena: A Realistic Web Environment for Building Autonomous Agents},
  author={Zhou, Shuyan and Xu, Frank F and Zhu, Hao and Zhou, Xuhui and Lo, Robert and Sridhar, Abishek and Cheng, Xianyi and Ou, Tianyue and Bisk, Yonatan and Fried, Daniel and others},
  booktitle={International Conference on Learning Representations},
  year={2024}
}

@InProceedings{zhang2025cu,
title = 	 {G-Designer: Architecting Multi-agent Communication Topologies via Graph Neural Networks},
author =       {Zhang, Guibin and Yue, Yanwei and Sun, Xiangguo and Wan, Guancheng and Yu, Miao and Fang, Junfeng and Wang, Kun and Chen, Tianlong and Cheng, Dawei},
booktitle = 	 {Proceedings of the 42nd International Conference on Machine Learning},
pages = 	 {76678--76692},
year = 	 {2025},
volume = 	 {267},
publisher =    {PMLR}
}

@article{mo2025multi,
  title={Multi-Agent Tool-Integrated Policy Optimization},
  author={Mo, Zhanfeng and Li, Xingxuan and Chen, Yuntao and Bing, Lidong},
  journal={arXiv preprint arXiv:2510.04678},
  year={2025}
}

@article{xu2025acesearcher,
  title={Acesearcher: Bootstrapping reasoning and search for llms via reinforced self-play},
  author={Xu, Ran and Zhuang, Yuchen and Dong, Zihan and Wang, Jonathan and Yu, Yue and Ho, Joyce C and Zhang, Linjun and Wang, Haoyu and Shi, Wenqi and Yang, Carl},
  journal={arXiv preprint arXiv:2509.24193},
  year={2025}
}

@article{grattafiori2024llama,
  title={The llama 3 herd of models},
  author={Grattafiori, Aaron and Dubey, Abhimanyu and Jauhri, Abhinav and Pandey, Abhinav and Kadian, Abhishek and Al-Dahle, Ahmad and Letman, Aiesha and Mathur, Akhil and Schelten, Alan and Vaughan, Alex and others},
  journal={arXiv preprint arXiv:2407.21783},
  year={2024}
}

@article{yang2025qwen3,
  title={Qwen3 technical report},
  author={Yang, An and Li, Anfeng and Yang, Baosong and Zhang, Beichen and Hui, Binyuan and Zheng, Bo and Yu, Bowen and Gao, Chang and Huang, Chengen and Lv, Chenxu and others},
  journal={arXiv preprint arXiv:2505.09388},
  year={2025}
}

@inproceedings{yu2023prompt,
  title={Prompt-Based Monte-Carlo Tree Search for Goal-oriented Dialogue Policy Planning},
  author={Yu, Xiao and Chen, Maximillian and Yu, Zhou},
  booktitle={Proceedings of the 2023 Conference on Empirical Methods in Natural Language Processing},
  pages={7101--7125},
  year={2023}
}

@article{jin2025search,
  title={Search-r1: Training llms to reason and leverage search engines with reinforcement learning},
  author={Jin, Bowen and Zeng, Hansi and Yue, Zhenrui and Yoon, Jinsung and Arik, Sercan and Wang, Dong and Zamani, Hamed and Han, Jiawei},
  journal={arXiv preprint arXiv:2503.09516},
  year={2025}
}

@article{shao2024deepseekmath,
  title={Deepseekmath: Pushing the limits of mathematical reasoning in open language models},
  author={Shao, Zhihong and Wang, Peiyi and Zhu, Qihao and Xu, Runxin and Song, Junxiao and Bi, Xiao and Zhang, Haowei and Zhang, Mingchuan and Li, YK and Wu, Yang and others},
  journal={arXiv preprint arXiv:2402.03300},
  year={2024}
}

@article{trivedi2022musique,
  title={MuSiQue: Multihop Questions via Single-hop Question Composition},
  author={Trivedi, Harsh and Balasubramanian, Niranjan and Khot, Tushar and Sabharwal, Ashish},
  journal={Transactions of the Association for Computational Linguistics},
  volume={10},
  pages={539--554},
  year={2022},
  publisher={MIT Press One Broadway, 12th Floor, Cambridge, Massachusetts 02142, USA~…}
}

@inproceedings{mialon2023gaia,
  title={Gaia: a benchmark for general ai assistants},
  author={Mialon, Gr{\'e}goire and Fourrier, Cl{\'e}mentine and Wolf, Thomas and LeCun, Yann and Scialom, Thomas},
  booktitle={The Twelfth International Conference on Learning Representations},
  year={2023}
}

@article{wu2025webwalker,
  title={Webwalker: Benchmarking llms in web traversal},
  author={Wu, Jialong and Yin, Wenbiao and Jiang, Yong and Wang, Zhenglin and Xi, Zekun and Fang, Runnan and Zhang, Linhai and He, Yulan and Zhou, Deyu and Xie, Pengjun and others},
  journal={arXiv preprint arXiv:2501.07572},
  year={2025}
}

@inproceedings{krishna2025fact,
  title={Fact, fetch, and reason: A unified evaluation of retrieval-augmented generation},
  author={Krishna, Satyapriya and Krishna, Kalpesh and Mohananey, Anhad and Schwarcz, Steven and Stambler, Adam and Upadhyay, Shyam and Faruqui, Manaal},
  booktitle={Proceedings of the 2025 Conference of the Nations of the Americas Chapter of the Association for Computational Linguistics: Human Language Technologies (Volume 1: Long Papers)},
  pages={4745--4759},
  year={2025}
}

@inproceedings{zhao2024docmath,
  title={DocMath-eval: Evaluating math reasoning capabilities of LLMs in understanding long and specialized documents},
  author={Zhao, Yilun and Long, Yitao and Liu, Hongjun and Kamoi, Ryo and Nan, Linyong and Chen, Lyuhao and Liu, Yixin and Tang, Xiangru and Zhang, Rui and Cohan, Arman},
  booktitle={Proceedings of the 62nd Annual Meeting of the Association for Computational Linguistics},
  pages={16103--16120},
  year={2024}
}

@article{napolitano2025lightningshap,
  title={LightningSHAP: A Cost-Effective Approach to Local Shapley Values Estimation},
  author={Napolitano, Davide and Cagliero, Luca},
  journal={IEEE Transactions on Artificial Intelligence},
  year={2025},
  publisher={IEEE}
}

@article{chen2023algorithms,
  title={Algorithms to estimate Shapley value feature attributions},
  author={Chen, Hugh and Covert, Ian C and Lundberg, Scott M and Lee, Su-In},
  journal={Nature Machine Intelligence},
  volume={5},
  number={6},
  pages={590--601},
  year={2023},
  publisher={Nature Publishing Group UK London}
}

@inproceedings{nauman2023many,
  title={On many-actions policy gradient},
  author={Nauman, Michal and Cygan, Marek},
  booktitle={International Conference on Machine Learning},
  pages={25769--25789},
  year={2023},
  organization={PMLR}
}

@article{ferrag2025llm,
  title={From llm reasoning to autonomous ai agents: A comprehensive review},
  author={Ferrag, Mohamed Amine and Tihanyi, Norbert and Debbah, Merouane},
  journal={arXiv preprint arXiv:2504.19678},
  year={2025}
}

@inproceedings{wang2024empowering,
  title={Empowering large language models: Tool learning for real-world interaction},
  author={Wang, Hongru and Qin, Yujia and Lin, Yankai and Pan, Jeff Z and Wong, Kam-Fai},
  booktitle={Proceedings of the 47th International ACM SIGIR Conference on Research and Development in Information Retrieval},
  pages={2983--2986},
  year={2024}
}

@article{zheng2025knowledge,
  title={Knowledge augmented complex problem solving with large language models: A survey},
  author={Zheng, Da and Du, Lun and Su, Junwei and Tian, Yuchen and Zhu, Yuqi and Zhang, Jintian and Wei, Lanning and Zhang, Ningyu and Chen, Huajun},
  journal={arXiv preprint arXiv:2505.03418},
  year={2025}
}

@inproceedings{tang2025finmmr,
  title={Finmmr: make financial numerical reasoning more multimodal, comprehensive, and challenging},
  author={Tang, Zichen and Liu, Jiacheng and Yang, Zhongjun and Li, Rongjin and Rong, Zihua and He, Haoyang and Hao, Zhuodi and Hu, Xinyang and Ji, Kun and Ma, Ziyan and others},
  booktitle={Proceedings of the IEEE/CVF International Conference on Computer Vision},
  pages={3245--3257},
  year={2025}
}

@article{yang2025spell,
  title={Spell: Self-play reinforcement learning for evolving long-context language models},
  author={Yang, Ziyi and Shen, Weizhou and Li, Chenliang and Chen, Ruijun and Wan, Fanqi and Yan, Ming and Quan, Xiaojun and Huang, Fei},
  journal={arXiv preprint arXiv:2509.23863},
  year={2025}
}

@article{chan2025infrastructure,
  title={Infrastructure for ai agents},
  author={Chan, Alan and Wei, Kevin and Huang, Sihao and Rajkumar, Nitarshan and Perrier, Elija and Lazar, Seth and Hadfield, Gillian K and Anderljung, Markus},
  journal={arXiv preprint arXiv:2501.10114},
  year={2025}
}

@inproceedings{zhang2025breaking,
  title={Breaking agents: Compromising autonomous llm agents through malfunction amplification},
  author={Zhang, Boyang and Tan, Yicong and Shen, Yun and Salem, Ahmed and Backes, Michael and Zannettou, Savvas and Zhang, Yang},
  booktitle={Proceedings of the 2025 Conference on Empirical Methods in Natural Language Processing},
  pages={34952--34964},
  year={2025}
}
\bibliographystyle{icml2026}


\clearpage
\onecolumn 

\appendix

\makeatletter
\let\oldsection\section
\let\oldsubsection\subsection

\renewcommand{\section}{\@ifstar\@starredsec\@unstarredsec}
\newcommand{\@starredsec}[1]{\oldsection*{#1}}
\newcommand{\@unstarredsec}[1]{%
\oldsection{#1}%
\addtocontents{toc}{\protect\contentsline{section}{\protect\numberline{\thesection}#1}{\thepage}{\@currentHref}}%
}

\renewcommand{\subsection}{\@ifstar\@starredsubsec\@unstarredsubsec}
\newcommand{\@starredsubsec}[1]{\oldsubsection*{#1}}
\newcommand{\@unstarredsubsec}[1]{%
\oldsubsection{#1}%
\addtocontents{toc}{\protect\contentsline{subsection}{\protect\numberline{\thesubsection}#1}{\thepage}{\@currentHref}}%
}
\makeatother

\clearpage
\begin{center}
\Large\bfseries Appendix
\end{center}
\vspace{1em}

\makeatletter
\renewcommand\l@section{\@dottedtocline{1}{0em}{2.5em}}
\renewcommand\l@subsection{\@dottedtocline{2}{1.5em}{3.2em}}

\let\old@dottedtocline\@dottedtocline
\def\@dottedtocline#1#2#3#4#5{%
\ifnum#1=1 
  \old@dottedtocline{#1}{#2}{#3}{\bfseries#4}{#5}%
\else
  \old@dottedtocline{#1}{#2}{#3}{#4}{#5}%
\fi
}

\hypersetup{linkcolor=black}
\setcounter{tocdepth}{1}
\@starttoc{toc}
\makeatother

\clearpage

\section{Algorithm Workflow}
\label{sec:appendix-algorithm}
\begin{algorithm}[!htbp]
\caption{Workflow of SHARP for Tool-Integrated Planning}
\label{alg:sharp_tip}
\DontPrintSemicolon
\footnotesize
\SetAlgoNlRelativeSize{-1}
\setlength{\algomargin}{2.5em}  
\SetNlSkip{0.6em}               
\SetInd{0.6em}{1.0em}           

\KwIn{
Query $q$; shared policy $\pi_{\theta}$; old policy $\pi_{\theta_{\mathrm{old}}}$;\;
role prompts $p_{\text{planner}}, p_{\text{worker}}$;\;
group size $G$; clip $\epsilon$; constant $\delta$;\;
reward weights $\alpha,\beta,\gamma$; planner scale $\lambda$;\;
tool reward $\phi(\cdot)$.
}
\KwOut{Updated parameters $\theta$.}

\BlankLine

\For{$i \leftarrow 1$ \KwTo $G$}{
\bluecomment{Collect TIP trajectories}\;

Sample planner trace $\tau_i^{0} \sim \pi_{\theta_{\mathrm{old}}}(\cdot \mid q, p_{\text{planner}})$\;
Obtain subtask actions $\{a_{i,t}\}_{t=1}^{T_i}$\;

\ForEach{subtask action $a_{i,t}$}{
Run a worker with $p_{\text{worker}}$, interact with tools, and record $\tau_i^{t}$\;
}

Form full trajectory $\tau_i = (\tau_i^{0}, \tau_i^{1}, \ldots, \tau_i^{T_i})$\;
Record participating worker set $\mathcal{M}_i$\;

\BlankLine
\bluecomment{Compute compositional rewards}\;

Compute broadcast reward $R_{\mathrm{acc}}(\tau_i) \in \{0,1\}$\;

\ForEach{$m \in \{0\}\cup\mathcal{M}_i$}{
$R^{\mathrm{b}}_{i,m} \leftarrow R_{\mathrm{acc}}(\tau_i)$\;
$R^{\mathrm{tool}}_{i,m} \leftarrow \sum_{j=1}^{T_{i,m}} \phi(a^{m}_{i,j}, s^{m}_{i,j})$\;
}

\BlankLine
\bluecomment{Counterfactual marginal credit (Shapley-style)}\;

\ForEach{$m \in \mathcal{M}_i$}{
Construct counterfactual $\tau_i^{\setminus m}$ by masking agent $m$\;
$\mathrm{credit}_{i,m} \leftarrow R_{\mathrm{acc}}(\tau_i) - R_{\mathrm{acc}}(\tau_i^{\setminus m})$\;
$R^{\mathrm{mc}}_{i,m} \leftarrow \mathrm{credit}_{i,m}$\;
}

$R^{\mathrm{mc}}_{i,0} \leftarrow \lambda \cdot \frac{1}{|\mathcal{M}_i|}
\sum_{m\in\mathcal{M}_i} \max(\mathrm{credit}_{i,m},0)$\;

\ForEach{$m \in \{0\}\cup\mathcal{M}_i$}{
$\bar{R}_{i,m} \leftarrow \alpha R^{\mathrm{b}}_{i,m}
+ \beta R^{\mathrm{mc}}_{i,m}
+ \gamma R^{\mathrm{tool}}_{i,m}$\;
}

\BlankLine
\bluecomment{Group-relative advantages and SHARP update}\;

}

\ForEach{agent identity $m$ in the sampled group}{
$\mu_m \leftarrow \mathrm{mean}(\{\bar{R}_{i,m}\})$, \;
$\sigma_m \leftarrow \mathrm{std}(\{\bar{R}_{i,m}\})$\;
}

\For{$i \leftarrow 1$ \KwTo $G$}{
\ForEach{$m \in \{0\}\cup\mathcal{M}_i$}{
$\hat{A}_{i,m} \leftarrow \dfrac{\bar{R}_{i,m}-\mu_m}{\sigma_m+\delta}$\;

$R_{i,m} \leftarrow \exp\!\Big(
\sum_{j=1}^{T_{i,m}} \log \pi_{\theta}(a^{m}_{i,j}\mid \mathrm{ctx}^{m}_{i,j})
-
\sum_{j=1}^{T_{i,m}} \log \pi_{\theta_{\mathrm{old}}}(a^{m}_{i,j}\mid \mathrm{ctx}^{m}_{i,j})
\Big)$\;

$R^{\mathrm{clip}}_{i,m} \leftarrow
\min\!\big(R_{i,m}\hat{A}_{i,m},\,
\mathrm{clip}(R_{i,m},1-\epsilon,1+\epsilon)\hat{A}_{i,m}\big)$\;
}
}

$J_{\mathrm{SHARP}}(\theta) \leftarrow
\frac{1}{G}\sum_{i=1}^{G}
\frac{1}{|\{0\}\cup\mathcal{M}_i|}
\sum_{m \in \{0\}\cup\mathcal{M}_i}
R^{\mathrm{clip}}_{i,m}$\;

Update $\theta \leftarrow \theta + \eta \nabla_{\theta} J_{\mathrm{SHARP}}(\theta)$\;

\end{algorithm}

\section{Information of Training and Test Dataset}
\label{sec:appendix-data}
\paragraph{Benchmarks.}
We consider five benchmarks covering multi-hop reasoning, tool-assisted interaction, and structured decision making:
\begin{itemize}[leftmargin=*, nosep]
  \item \textbf{MuSiQue} \cite{trivedi2022musique}: a multi-hop question answering benchmark constructed by composing connected single-hop questions, designed to require genuine multi-step reasoning.
  \item \textbf{GAIA-text} \cite{mialon2023gaia}: evaluates general AI assistants on real-world questions requiring reasoning, web browsing, and tool-use proficiency.
  \item \textbf{WebWalkerQA} \cite{wu2025webwalker}: benchmarks multi-step web traversal and evidence aggregation through structured navigation.
  \item \textbf{FRAMES} \cite{krishna2025fact}: a unified benchmark for evaluating factuality, retrieval, and reasoning in end-to-end retrieval-augmented generation.
  \item \textbf{DocMath-Eval} \cite{zhao2024docmath}: evaluates document-grounded numerical reasoning over long-context documents with text and tables.
\end{itemize}

\paragraph{Training and Test Splits.}
We use a subset of these benchmarks for training and evaluation.
For MuSiQue, we train our models on 5{,}975 instances and evaluate on a held-out test set of 200 instances.
For GAIA-text, WebWalkerQA, FRAMES, and DocMath-Eval, we do not use any training data and evaluate our models directly on their released test sets following the official evaluation protocols.

\paragraph{Data Release.}
To ensure reproducibility and facilitate future research, we will release all training and test splits used in our experiments, together with preprocessing scripts and evaluation protocols, upon publication.

\section{Implementation Details}
\label{sec:appendix-impl}
As stated in the main manuscript, SHARP outperforms both single-agent and multi-agent baselines, achieving empirical gains of 23.66\% and 14.05\%, as confirmed by comparisons with baselines built on the Qwen3-8B backbones for fairness. The involved baselines include Plan-Search, Single-agent GRPO, Planner–Worker, CARD, COA and MATPO. Those models designed with LLaMA-3.1-8B or Qwen2.5-7B, or purely typology strategy (G-Designer) are not involved for fairness.

\paragraph{Training Setup.}
Our optimization framework leverages pre-trained weights from the \emph{Qwen-3-8B} and \emph{LLaMA-3.1-8B} instruction-tuned models. We employ the AdamW optimizer with a fixed learning rate of $10^{-5}$ and a global batch size of 256. To enhance trajectory diversity during the learning phase, we generate 8 rollouts per input instance. The training process is executed over 180 gradient steps across a cluster of 64$\times$A100 GPUs, using data parallelism to maintain throughput. Hyperparameters remain consistent across all benchmark evaluations unless otherwise noted, with reward coefficients specifically tuned to $\alpha=0.9$, $\beta=0.9$, and $\gamma=0.1$.

\paragraph{Agent Configuration.}
We implement a Multi-Agent System (MAS) predicated on role specialization. The architecture distributes task execution between a high-level Planner and one or more Workers ($K \ge 1$). While all agents are powered by the same underlying policy $\pi_\theta$, their functional behaviors are bifurcated via role-specific prompting. Specifically, the Planner is responsible for decomposing complex objectives into a discrete sequence of subtasks. The Worker agents execute these subtasks iteratively through direct interaction with the environment and tools.

\paragraph{SHARP Optimization.}
To facilitate fine-grained attribution, we utilize the SHARP optimization protocol. For each generated trajectory $\tau_i$, we perform counterfactual analysis by systematically masking the contributions of agent $m$ to produce a modified trajectory $\tau_i^{\setminus m}$. The marginal credit for each agent is subsequently derived from the delta in accumulated reward:
$$\mathrm{credit}_{i,m} = R_{\mathrm{acc}}(\tau_i) - R_{\mathrm{acc}}(\tau_i^{\setminus m})$$

We estimate group-wise normalization statistics ($\mu_m, \sigma_m$) across each mini-batch, maintaining stability with parameters $\epsilon=0.2$ and $\delta=10^{-6}$ throughout all experimental runs.

\paragraph{Tool Execution.}
Environment interactions are mediated through a role-scoped service interface built on FastMCP. This architecture enforces a strict separation of concerns, where the primary agent manages delegation and offloads execution to specialized auxiliary agents with non-overlapping toolsets.
For the computational sandbox, A dedicated Python agent connects to a remote environment via the python-mcp-server. It exposes an \emph{execute\_python\_code} tool that allows the safe execution of dynamic code with configurable timeouts.
Moreover, for the object of web exploration, the Browsing agent interfaces with a \emph{duck-mcp-server}, providing a sequential pipeline of \emph{duckduckgo\_search} and scrape for real-time information retrieval.

To ensure reproducibility and operational safety, tool access is strictly governed by role-scoping. Primary planner agents are limited to delegation tools, whereas execution agents interact solely with their designated backends. Furthermore, each agent is constrained to \emph{a single tool call per step}, preventing uncontrolled execution cycles.

\section{Cost Analysis}
\label{sec:cost-analysis}
\label{app:cost-analysis}

\begin{figure}[t]
\centering
\includegraphics[width=0.9\linewidth]{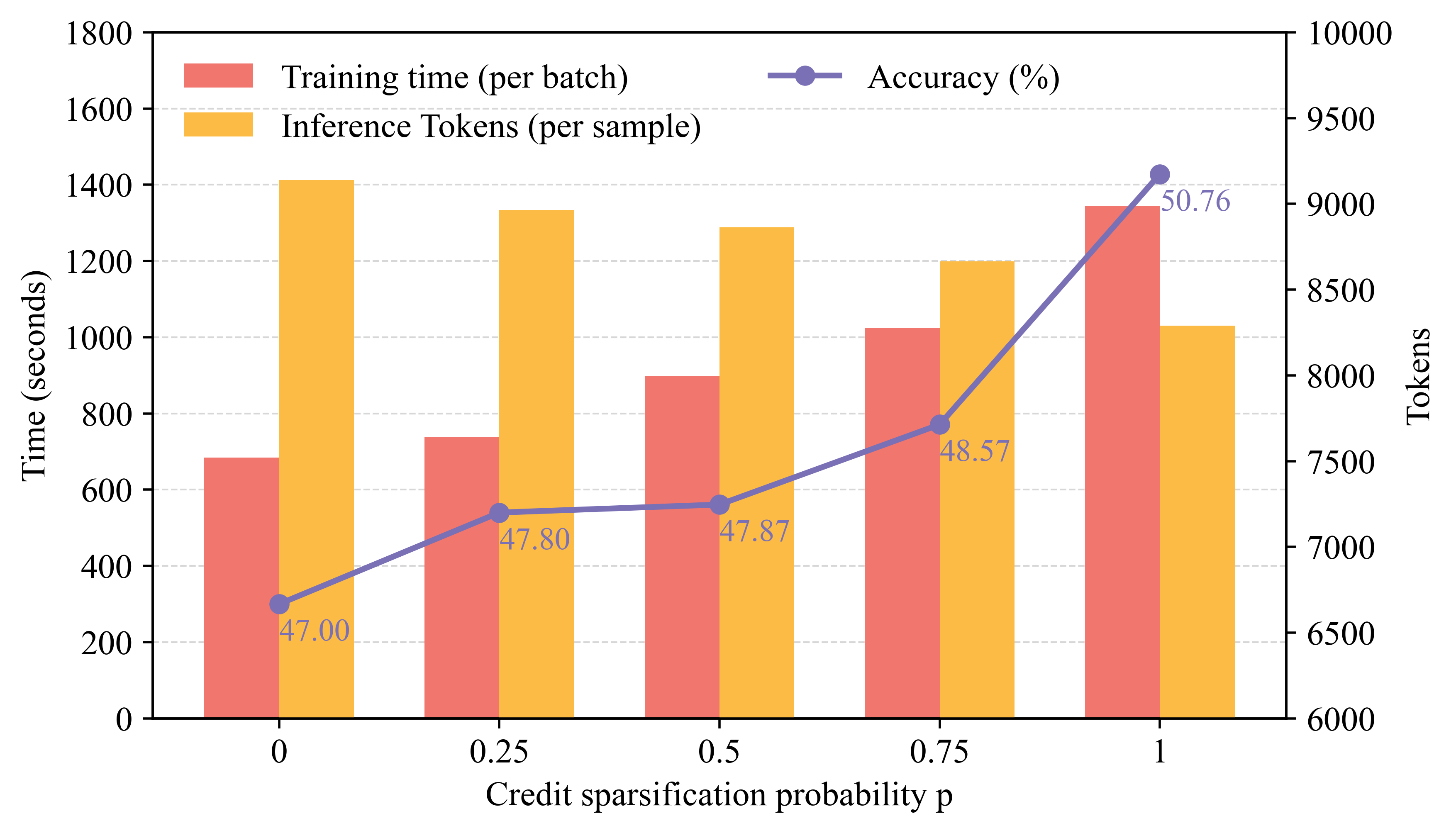}
\caption{\textbf{Analysis of the cost v.s. performance trade-off across varying credit sparsification levels.}
We adjust the credit sparsification probability $p$, where a $p$-fraction of subagent invocations undergo Shapley-based marginal credit assignment, while the remaining $(1-p)$ are assigned zero marginal credit.
The left axes and bars represent per-batch training latency, while the right axes and bars denote per-sample inference token consumption; the line plot indicates the corresponding task accuracy.
The results demonstrate that while increasing $p$ necessitates higher training expenditure, it simultaneously reduces inference-time token usage and enhances task performance.
This trend suggests that denser Shapley signals during training foster more efficient and precise coordination during deployment.}
\label{fig:cost_tradeoff}
\vspace{-0.5cm}
\end{figure}

\paragraph{Setup.}
To evaluate the impact of Shapley-based marginal credit on both training overhead and inference efficiency, we introduce a sparsification parameter $p \in [0,1]$. Specifically, for any given trajectory, Shapley-based credit is computed for only a $p$-fraction of the participating subagents, whereas the remaining $(1-p)$ fraction is assigned a null marginal credit without triggering the Shapley computation.
This control mechanism allows us to analyze the trade-off between the density of the credit signal and the associated computational burden during the training phase.

\paragraph{Empirical Trade-off and Performance Gains.}
As illustrated in Figure~\ref{fig:cost_tradeoff}, the system exhibits a clear shift in computational expenditure from inference to training as $p$ increases.
The training latency scales with $p$ due to the increased frequency of counterfactual evaluations required for marginal-credit estimation, with per-batch time rising from approximately $684$s at $p=0$ to $1345$s at $p=1$.
However, this investment in training yields substantial dividends in deployment efficiency: inference-time token consumption consistently declines, from $\sim$9.1k to $\sim$8.3k. Crucially, this gain in efficiency is accompanied by a performance boost, with task accuracy improving from $47.00\%$ to $50.76\%$.
In essence, the additional training cost facilitates the discovery of more concise and effective coordination strategies.

\paragraph{Mechanistic Insight: Why Credit Matters.}
The inverse relationship between training cost and inference token usage stems from the enhanced causal attribution provided by marginal credit.
Standard broadcast rewards often provide a noisy or ambiguous signal, which may inadvertently encourage agents to generate redundant intermediate steps or over-utilize tools due to poorly localized responsibility.
In contrast, Shapley-based credit provides a more rigorous approximation of each subagent's actual contribution to the final outcome.
By training more agents under such targeted feedback (higher $p$), the policy learns to suppress low-utility invocations and minimize redundant actions, ultimately resulting in shorter, more purposeful inference trajectories.

\paragraph{Computational Tractability in Practice.}
While Shapley estimation introduces additional computation, its impact on the overall training pipeline remains manageable for two primary reasons. First, marginal credit evaluations for different subagents are independent and can be parallelized, so the wall-clock time does not grow linearly with the number of agents.
Second, these computations are confined to trajectory-level counterfactual evaluation (i.e., masking specific subagent) and do not require rerunning the full optimization process.

\paragraph{Takeaway.}
The result in Figure~\ref{fig:cost_tradeoff} reinforces the concept of Shapley-based credit as a \emph{compute-for-coordination} mechanism. By allocating more computational resources to credit assignment during training, we can derive a more selective and intelligent multi-agent policy. This approach successfully trades a modest increase in offline training time for significantly more efficient and accurate autonomous execution in production environments.

\section{Catastrophic Forgetting Analysis}
\label{sec:forget-analysis}

\label{app:forget-analysis}

\paragraph{Setup.}
A common concern when training on a specialized task distribution (e.g., search/tool-heavy trajectories)
is \emph{catastrophic forgetting} of general domain knowledge.
To assess the impact of search-domain optimization on the model’s underlying knowledge base, we evaluate the Qwen3 backbone on the MMLU benchmark across three distinct configurations: 
(i) \textbf{Planner--Worker (Baseline)}, an inference-only hierarchical setup without further training; 
(ii) \textbf{MATPO}, representing a standard policy optimization approach trained on the search dataset; and (iii) \textbf{SHARP}, our proposed framework utilizing Shapley-based marginal credit on the same training distribution.

\begin{table}[t]
\centering
\small
\caption{\textbf{Assessment of Knowledge Retention via MMLU.} We report MMLU accuracies to monitor potential catastrophic forgetting following domain-specific training. Both MATPO and SHARP demonstrate that specialized optimization does not degrade general-domain knowledge. Instead, the trained variants exhibit marginal performance gains over the inference-only baseline, with SHARP achieving the highest overall accuracy.}
\label{tab:forgetting_mmlu}
\begin{tabular}{l c}
\toprule
\textbf{Method} & \textbf{MMLU (Accuracy $\uparrow$)} \\
\midrule
Planner--Worker (no train) & 93.04 \\
MATPO (search-trained)     & 94.00 \\
sharp (search-trained)      & 94.65 \\
\bottomrule
\end{tabular}
\end{table}

\paragraph{Results and discussion.}
As detailed in Table~\ref{tab:forgetting_mmlu}, specialized training on search-centric trajectories does not lead to a regression in general-domain performance. On the contrary, both trained models surpass the inference-only Planner--Worker baseline (93.04\%). MATPO shows a modest improvement (94.00\%), while SHARP achieves the most significant gain (94.65\%). These results provide empirical evidence that our training protocol preserves, rather than erodes, the pre-trained knowledge base. The observed performance uplift suggests a degree of \emph{positive transfer}, where the refined reasoning and coordination capabilities acquired during search-domain training generalize to improve zero-shot knowledge retrieval.

\paragraph{Why forgetting is mitigated?}
We hypothesize that the alleviation of catastrophic forgetting is rooted in the nature of the SHARP optimization objective. Unlike standard fine-tuning that might necessitate substantial representation shifts to accommodate new factual content, SHARP primarily refines \emph{agent coordination and credit assignment} logic. In broadcast-based reinforcement learning, the global reward signal is often confounded, potentially leading to noisy and indiscriminate parameter updates across the model. In contrast, SHARP utilizes Shapley-based marginal credit to provide a high-fidelity, targeted signal that localizes the causal contribution of specific subagent actions. By improving the signal-to-noise ratio of the updates, SHARP minimizes unnecessary perturbations to the model's latent representations, thereby safeguarding its general-purpose knowledge. Collectively, these findings underscore the robustness of our framework in specialized environments without sacrificing the foundational capabilities of the backbone LLM.

\section{Case Study: Raw Agent Trajectories}
\label{sec:appendix-case}

\newtcolorbox{fullbox}{
colback=blue!5,
colframe=blue!60!black,
title=\textbf{Full Response Trajectory},
breakable,
sharp corners,
listing only,
listing options={
basicstyle=\ttfamily\footnotesize,
breaklines=true,
breakatwhitespace=true,
columns=fullflexible
}
}

\newtcolorbox{cfbox}{
colback=red!5,
colframe=red!60!black,
title=\textbf{Counterfactual Response Trajectory},
breakable,
sharp corners,
listing only,
listing options={
basicstyle=\ttfamily\footnotesize,
breaklines=true,
breakatwhitespace=true,
columns=fullflexible
}
}

\newtcolorbox{subbox}{
colback=green!5,
colframe=green!60!black,
title=\textbf{Summarized Sub-agent Trajectory},
breakable,
sharp corners,
listing only,
listing options={
basicstyle=\ttfamily\footnotesize,
breaklines=true,
breakatwhitespace=true,
columns=fullflexible
}
}

\newtcolorbox{probbox}{
colback=gray!5,
colframe=black!60,
title=\textbf{Problem Statement},
breakable,
sharp corners,
listing only,
listing options={
basicstyle=\ttfamily\footnotesize,
breaklines=true,
breakatwhitespace=true,
columns=fullflexible
}
}


This section presents \textbf{verbatim agent trajectories}.
All contents are preserved exactly as generated by the agents,
without any summarization, rewriting, or post-hoc interpretation.

\subsection{Case Study 1: Useful Sub-agent}

\begin{probbox}
By 1931, what percentage of the world's Jews were the group of Jews that represented the bulk of modern Jewry?
\end{probbox}

\begin{fullbox}
\medskip

\texttt{<think>}
Okay, the user is asking about the percentage of the world's Jews in 1931 that belonged to the group representing the bulk of modern Jewry. First, I need to understand what group that is. Modern Jewry is mostly made up of Ashkenazi Jews, right? They are of Central and Eastern European origin. But I'm not 100\% sure, so I should verify that.

Wait, the question is about the percentage in 1931. So I need to find out the distribution of Jewish populations by region in that year. The main regions were Europe, the US, and others. Ashkenazi Jews were predominant in Europe, but by 1931, there might have been significant migration, especially due to the rise of antisemitism and the start of the Holocaust. Maybe the percentage had shifted?

I should check historical data on Jewish populations in 1931. The tool provided is search\_and\_browse from the browsing\_agent server. I need to use that to find the specific percentage. The subtask should be to find the percentage of world Jews in 1931 that were part of the dominant group (Ashkenazi).

But wait, maybe the user is referring to the group that became the majority in modern times, which might not be the same as the 1931 distribution. Let me make sure. Modern Jewry is largely Ashkenazi, but the question is about the percentage in 1931. So the subtask is to find that percentage.

I should use the search\_and\_browse tool with the subtask: ``What percentage of the world's Jews in 1931 were Ashkenazi Jews, representing the bulk of modern Jewry?'' Then, the tool will return the answer. Once I get that, I can present it as the final answer.
\texttt{</think>}

\medskip

\texttt{<use\_mcp\_tool>}\\
\texttt{<server\_name>browsing\_agent</server\_name>}\\
\texttt{<tool\_name>search\_and\_browse</tool\_name>}\\
\texttt{<arguments>}\\
\texttt{\{}\\
\texttt{\ \ ``subtask'': ``What percentage of the world's Jews in 1931 were Ashkenazi Jews, representing the bulk of modern Jewry?''}\\
\texttt{\}}\\
\texttt{</arguments>}\\
\texttt{</use\_mcp\_tool>}

\begin{subbox}
\medskip

\textbf{Final Response: Percentage of World Jews Identified as Ashkenazi in 1931}

\medskip

\textbf{Conclusion:}

By 1931, \textbf{92\%} of the world's Jews were Ashkenazi Jews, representing the bulk of modern Jewry. This percentage is widely cited in historical records and aligns with the peak demographic dominance of Ashkenazi Jews prior to the Holocaust.

\medskip

\textbf{Supporting Information:}
\begin{itemize}
\item \textbf{New World Encyclopedia}: Explicitly states, ``By 1931, Ashkenazi Jews accounted for nearly 92 percent of world Jewry.''
\item \textbf{LinkedIn Post}: References ``92\% of the world's Jews'' as the figure for Ashkenazi Jews in 1931, contextualized within historical demographic trends.
\item \textbf{Historical Forum Discussion}: Mentions that ``at their peak in 1931, Ashkenazi Jews accounted for 92 percent of the world's Jews,'' emphasizing this as a significant historical threshold.
\end{itemize}

\medskip

\textbf{Observations:}
\begin{itemize}
\item The figure of 92\% is consistently repeated across multiple independent sources, suggesting a high degree of consensus in historical scholarship.
\item The percentage reflects the pre-Holocaust demographic distribution, which is critical for understanding the demographic shifts that followed.
\item Minor variations in statistical interpretation exist, but the 92\% figure is the most frequently referenced and reliable estimate.
\end{itemize}

\medskip

\textbf{Contribution to Main Task:}

This subtask directly answers the main task by identifying Ashkenazi Jews as the group constituting \textbf{92\%} of the world's Jews in 1931. This information provides the foundational demographic data required to address the broader question. The main agent may now proceed to analyze post-1931 demographic changes or explore the historical significance of this percentage in shaping modern Jewry.
\end{subbox}

Answer:
\[
\boxed{92\%}
\]
\end{fullbox}

\begin{cfbox}

\medskip

\texttt{<think>}
Okay, let me try to figure this out. The question is asking about the percentage of the world's Jews in 1931 that were Ashkenazi Jews, who represent the bulk of modern Jewry.

First, I know that Ashkenazi Jews are a major group within the Jewish population, traditionally from Central and Eastern Europe. Before the Holocaust, they made up a significant portion of the global Jewish population. However, I need to find the exact percentage for the year 1931.

I remember that historical data shows that Ashkenazi Jews were the largest group, often around 80--90\% of the total Jewish population in the early twentieth century. But I should verify this. Since the user mentioned using the tool, but the tool's response was masked, I have to rely on known historical facts.

In 1931, before the Holocaust, the majority of Jews were indeed Ashkenazi. The percentage is typically cited as around 80\% to 90\%, but I need to be precise. Some sources state that by the early 1900s, Ashkenazi Jews constituted about 85\% of the world's Jewish population. However, exact figures can vary slightly depending on the source.

Given that the user wants the answer in a box, I should provide the most commonly accepted figure. Based on historical records, the percentage is approximately 85\%. Therefore, the answer is 85\%.
\texttt{</think>}

\medskip

The group of Jews that represented the bulk of modern Jewry by 1931 was the Ashkenazi Jews. Historical data indicates that Ashkenazi Jews accounted for approximately \textbf{85\%} of the world's Jewish population during this period.

Answer:
\[
\boxed{85}
\]

\end{cfbox}

\paragraph{Example}
This case illustrates the practical usefulness of sub-agents in our multi-agent framework.
When sub-agents are enabled, the system produces the correct answer: Ashkenazi Jews constituted approximately 92\% of the world’s Jewish population by 1931. 
The sub-agent actively retrieves historical evidence and verifies the numerical figure using multiple sources, leading to a precise and well-grounded final response.

In contrast, the counterfactual setting without sub-agents fails to reach the correct conclusion.
Although the main agent correctly recognizes Ashkenazi Jews as the dominant group, it relies on coarse prior knowledge and outputs an approximate but incorrect estimate (85\%).
This reflects insufficient evidence grounding and limited numerical precision.

The comparison clearly shows that sub-agents contribute useful and non-redundant information.
By performing targeted retrieval and cross-checking, the sub-agent directly resolves the main agent’s uncertainty and changes the outcome from incorrect to correct.
This case demonstrates that sub-agents improve reasoning accuracy rather than merely increasing computational cost.

\subsection{Case Study 2: Harmful Sub-agent}

\begin{probbox}
Which party held the 1781 governorship of Virginia?
\end{probbox}

\begin{fullbox}

\texttt{<think>}

The user is asking which party held the governorship of Virginia in 1781.  
To approach this question, it is first necessary to identify who served as the
governor of Virginia during that year.

Because this specific historical information may not be immediately available
from memory, the appropriate strategy is to consult an external information
retrieval tool. The available tool for this purpose is
\texttt{search\_and\_browse} provided by the \texttt{browsing\_agent} server.

The immediate subtask is therefore to identify the governor of Virginia in 1781.
The tool requires a clearly specified subtask in structured form.
Accordingly, the query should be framed to retrieve not only the governor's
identity but also their political affiliation.

However, the user explicitly asks about the \emph{political party}, not merely
the individual officeholder. This introduces an additional complication, as
formal political parties may not yet have existed during the period in question.
Nevertheless, the subtask can be formulated as:

\emph{``Identify the political party of the governor of Virginia in 1781.''}

If the tool returns only the governor's name, a secondary reasoning step can be
used to infer political alignment from historical context. The process therefore
proceeds by issuing this subtask and extracting the relevant information from
the result.

\texttt{</think>}

\medskip

\texttt{<use\_mcp\_tool>}\\
\texttt{<server\_name>browsing\_agent</server\_name>}\\
\texttt{<tool\_name>search\_and\_browse</tool\_name>}\\
\texttt{<arguments>}\\
\texttt{\{}\\
\texttt{\ \ ``subtask'': ``Identify the political party of the governor of Virginia in 1781.''}\\
\texttt{\}}\\
\texttt{</arguments>}\\
\texttt{</use\_mcp\_tool>}

\begin{subbox}

\medskip

\textbf{Final Response: Political Party of the 1781 Virginia Governor}

\medskip

\textbf{Conclusion.}  
The governor of Virginia in 1781 was \textbf{Thomas Jefferson}. He was associated
with Anti-Federalist and early Republican political principles. Importantly,
no formal political party structure existed in Virginia or the United States at
that time.

\medskip

\textbf{Supporting Information.}
\begin{itemize}
  \item \textbf{Thomas Jefferson's role:} Served as governor of Virginia from
  1779 to 1781 during the American Revolutionary War.
  \item \textbf{Political context:} Formal political parties such as the
  Democratic--Republican Party and the Federalist Party emerged later, primarily
  in the 1790s.
  \item \textbf{Factional alignment:} Jefferson opposed a strong centralized
  federal government, a position later associated with the
  Democratic--Republican tradition.
\end{itemize}

\medskip

\textbf{Observations.}
\begin{itemize}
  \item The question applies a modern concept of political parties to a period
  that predates formal party systems.
  \item Jefferson's later party affiliations should not be retroactively applied
  to his 1781 governorship.
  \item Historical records from this period typically describe political
  alignments in terms of factions rather than parties.
\end{itemize}

\medskip

\textbf{Contribution to Main Task.}  
This analysis establishes that the governorship of Virginia in 1781 was not held
by any formal political party. Instead, it was occupied by an individual aligned
with Anti-Federalist and early Republican ideological currents. This clarification
allows the main task to correctly conclude that the notion of a ruling party in
1781 is historically anachronistic.
\end{subbox}

\medskip

\[
\boxed{\text{No formal party; associated with Anti-Federalist/Republican factions}}
\]

\end{fullbox}

\begin{cfbox}

\texttt{<think>}

The question to be addressed is: \emph{``Which party held the 1781 governorship of
Virginia?''}

To answer this, it is necessary to consider the political landscape of Virginia
in the late eighteenth century. During this period, the United States was still
in its formative years, and political parties were not yet formalized in the
modern sense. Political alignment was primarily ideological rather than
institutional.

The two dominant ideological groupings that later evolved into formal parties
were the Federalists and the Democratic--Republicans. However, these factions
were loosely organized and did not operate as structured parties with formal
membership or official control of offices.

In 1781, the governor of Virginia was Thomas Jefferson. Jefferson was a prominent
leader of the American Revolutionary period and a strong advocate of states'
rights, limited central authority, and individual liberties. These views later
became closely associated with the Democratic--Republican tradition.

Although Jefferson is historically linked to the Democratic--Republican Party,
it is important to emphasize that this party did not yet exist as a formal
institution in 1781. Political identity at the time was fluid, and officeholders
were not elected under formal party labels.

Nevertheless, if the question is interpreted through a retrospective
party-alignment lens, Jefferson's ideological position aligns most closely with
the Democratic--Republican faction. On this basis, one may attribute the 1781
governorship to that ideological grouping, while recognizing the historical
limitations of such a classification.

\texttt{</think>}

\medskip

The governor of Virginia in 1781 was Thomas Jefferson. His political views and
later affiliations place him within the Democratic--Republican ideological
tradition. Accordingly, when interpreted through a retrospective party framework,
the governorship may be associated with the Democratic--Republican faction.

\medskip

\[
\boxed{\text{Democratic--Republican}}
\]

\end{cfbox}

\paragraph{Example}
This case highlights a failure mode where a sub-agent becomes \emph{harmful} and pushes the system toward an incorrect final answer \cite{zhang2025breaking,chan2025infrastructure}.
The question asks: \emph{“Which party held the 1781 governorship of Virginia?”}
In the \textbf{counterfactual} trajectory \emph{without} sub-agents, the main agent answers \textbf{Democratic--Republican}, which matches the expected label in the benchmark.

However, in the \textbf{full} trajectory \emph{with} a sub-agent, the final answer is diverted to
\textbf{“No formal party; associated with Anti-Federalist/Republican factions”}, which is marked incorrect by the task.
Although the sub-agent’s reasoning is historically nuanced (formal parties were not fully institutionalized in 1781), it introduces an anachronism caveat and shifts the response away from the dataset’s labeling scheme.
As a result, the sub-agent injects \emph{unhelpful} complexity and overrides the simpler, benchmark-aligned answer, turning a correct outcome into an incorrect one.

This contrast shows that sub-agents are not always beneficial: when they over-emphasize side conditions or redefine the question, they can \emph{harm} performance.
It motivates the need for credit assignment (and training signals) that penalize sub-agents whose contributions reduce task success, rather than rewarding all additional computation equally.

\clearpage\newpage
\section{System Prompts: Planner Agent, and Worker Agents}
\definecolor{promptblue}{RGB}{0,51,153}   

\newtcolorbox{promptbox}[2][]{%
  enhanced,
  breakable, 
  colback=white,
  colframe=promptblue,
  boxrule=1.2pt,
  arc=2mm,
  outer arc=2mm,
  top=3mm,
  bottom=3mm,
  left=3mm,
  right=3mm,
  fonttitle=\bfseries\color{white},
  coltitle=white,
  title=#2,
  attach boxed title to top left={
    xshift=2mm,
    yshift*=-\tcboxedtitleheight/2
  },
  boxed title style={
    colback=promptblue,
    boxrule=0pt,
    arc=2mm,
    left=4mm,
    right=4mm,
    top=1.5mm,
    bottom=1.5mm
  },
  listing only,
  listing options={
    basicstyle=\ttfamily\footnotesize,
    breaklines=true,
    showstringspaces=false,
    columns=fullflexible,
    keepspaces=true
  },
  #1
}

\footnotesize
\begin{promptbox}{System Prompt and Tool Schema of the Planner Agent}
\ttfamily

In this environment you have access to a set of tools you can use to answer the user's question.

You only have access to the tools provided below. You can only use one tool per message, and will receive the result of that tool in the user's next response. You use tools step-by-step to accomplish a given task, with each tool-use informed by the result of the previous tool-use. Today is: 2025-07-16.

\medskip
\textbf{Tool-Use Formatting Instructions}

Tool-use is formatted using XML-style tags. The tool-use is enclosed in \texttt{<use\_mcp\_tool>} and \texttt{</use\_mcp\_tool>} and each parameter is similarly enclosed within its own set of tags.

The Model Context Protocol (MCP) connects to servers that provide additional tools and resources to extend your capabilities. You can use the server's tools via the \texttt{use\_mcp\_tool} interface.

\medskip
\textbf{Description}

Request to use a tool provided by an MCP server. Each MCP server can provide multiple tools with different capabilities. Tools have defined input schemas that specify required and optional parameters.

\medskip
\textbf{Parameters}

\begin{itemize}
  \item \texttt{server\_name}: (required) The name of the MCP server providing the tool.
  \item \texttt{tool\_name}: (required) The name of the tool to execute.
  \item \texttt{arguments}: (required) A JSON object containing the tool's input parameters, following the tool's input schema.
\end{itemize}

\medskip
\textbf{Usage}

\begin{verbatim}
<use_mcp_tool>
  <server_name>server_name_here</server_name>
  <tool_name>tool_name_here</tool_name>
  <arguments>
  {
    "param1": "value1",
    "param2": "value2 \"escaped string\""
  }
  </arguments>
</use_mcp_tool>
\end{verbatim}

\medskip
\textbf{Important Notes}

\begin{itemize}
  \item Tool-use must be placed at the end of your response, at the top level, and not nested within other tags.
  \item Always adhere to this format for tool use to ensure proper parsing and execution.
  \item String and scalar parameters should be specified as is, while lists and objects should use JSON format.
  \item The output is not expected to be valid XML and is parsed with regular expressions.
\end{itemize}

\medskip
\textbf{Available Functions}

\textbf{Server name:} browsing\_agent

\textbf{Tool name:} search\_and\_browse

This tool performs the subtask of searching and browsing the web for specific missing information and generating the desired answer. The subtask should be clearly defined, include relevant background, and focus on factual gaps.

\medskip
\textbf{Input JSON Schema}

\begin{verbatim}
{
  "properties": {
    "subtask": {
      "title": "Subtask",
      "type": "string"
    }
  },
  "required": ["subtask"],
  "title": "search_and_browseArguments",
  "type": "object"
}
\end{verbatim}

\medskip
\textbf{General Objective}

You accomplish a given task iteratively, breaking it down into clear steps and working through them methodically.

\medskip
\textbf{Task Strategy}

\begin{enumerate}
  \item Analyze the user's request and set clear, achievable sub-goals.
  \item Start with a concise, numbered, step-by-step plan before taking any action.
  \item Work through these sub-goals sequentially and adjust the plan as needed.
  \item Use tools strategically to accomplish each sub-goal.
  \item Revise earlier steps if new information emerges.
\end{enumerate}

\medskip
\textbf{Tool-Use Guidelines}

\begin{enumerate}
  \item Each step must involve a single tool call, unless the task is already solved.
  \item Before each tool call:
  \begin{itemize}
    \item Summarize what is known.
    \item Identify what is missing.
    \item Choose the most relevant tool.
    \item Verify all required parameters.
  \end{itemize}
  \item All tool queries must include full context.
  \item Avoid vague queries. Each call should retrieve actionable information.
  \item Extract and summarize partial information if a tool result is incomplete.
\end{enumerate}

\medskip
\textbf{Tool-Use Communication Rules}

\begin{enumerate}
  \item Do not include tool results in your response.
  \item Do not present the final answer until the entire task is complete.
  \item Do not mention tool names.
  \item Do not engage in unnecessary back-and-forth.
  \item Do not use non-existent tools.
  \item Respond in the same language as the user's message.
  \item If the task does not require tool use, answer directly.
\end{enumerate}

\medskip
\textbf{Agent Specific Objective}

You are a task-solving agent that uses tools step-by-step to answer the user's question. Your goal is to provide complete, accurate, and well-reasoned answers using additional tools.

\end{promptbox}
\normalsize

\footnotesize
\begin{promptbox}{System Prompt and Tool Schema of the Worker Agent}
\ttfamily

In this environment you have access to a set of tools you can use to answer the user's question.

You only have access to the tools provided below. You can only use one tool per message, and will receive the result of that tool in the user's next response. You use tools step-by-step to accomplish a given task, with each tool-use informed by the result of the previous tool-use. Today is: 2025-07-16.

\medskip
\textbf{Tool-Use Formatting Instructions}

Tool-use is formatted using XML-style tags. The tool-use is enclosed in \texttt{<use\_mcp\_tool>} and \texttt{</use\_mcp\_tool>}, and each parameter is similarly enclosed within its own set of tags.

The Model Context Protocol (MCP) connects to servers that provide additional tools and resources to extend your capabilities. You can use the server's tools via the \texttt{use\_mcp\_tool} interface.

\medskip
\textbf{Description}

Request to use a tool provided by an MCP server. Each MCP server can provide multiple tools with different capabilities. Tools have defined input schemas that specify required and optional parameters.

\medskip
\textbf{Parameters}

\begin{itemize}
  \item \texttt{server\_name}: (required) The name of the MCP server providing the tool.
  \item \texttt{tool\_name}: (required) The name of the tool to execute.
  \item \texttt{arguments}: (required) A JSON object containing the tool's input parameters, following the tool's input schema.
\end{itemize}

\medskip
\textbf{Usage}

\begin{verbatim}
<use_mcp_tool>
  <server_name>server_name_here</server_name>
  <tool_name>tool_name_here</tool_name>
  <arguments>
  {
    "param1": "value1",
    "param2": "value2 \"escaped string\""
  }
  </arguments>
</use_mcp_tool>
\end{verbatim}

\medskip
\textbf{Important Notes}

\begin{itemize}
  \item Tool-use must be placed at the end of your response, at the top level, and not nested within other tags.
  \item Always adhere to this format for tool use to ensure proper parsing and execution.
  \item String and scalar parameters should be specified as is, while lists and objects should use JSON format.
  \item The output is not expected to be valid XML and is parsed with regular expressions.
\end{itemize}

\medskip
\textbf{Available Functions}

\textbf{Server name:} search\_and\_scrape

\medskip
\textbf{Tool name:} search

This tool searches the web for relevant sources given a query and returns a list of results (e.g., title, snippet, and URL). Use it to locate candidate sources before scraping.

\medskip
\textbf{Input JSON Schema (search)}

\begin{verbatim}
{
  "properties": {
    "query": {
      "title": "Query",
      "type": "string"
    },
    "recency_days": {
      "title": "RecencyDays",
      "type": "integer"
    },
    "domains": {
      "title": "Domains",
      "type": "array",
      "items": { "type": "string" }
    }
  },
  "required": ["query"],
  "title": "searchArguments",
  "type": "object"
}
\end{verbatim}

\medskip
\textbf{Tool name:} scrape

This tool fetches and extracts the content from a given URL. Use it to retrieve the full text needed for answering and for quoting/verifying claims.

\medskip
\textbf{Input JSON Schema (scrape)}

\begin{verbatim}
{
  "properties": {
    "url": {
      "title": "URL",
      "type": "string"
    }
  },
  "required": ["url"],
  "title": "scrapeArguments",
  "type": "object"
}
\end{verbatim}

\medskip
\textbf{General Objective}

You accomplish a given task iteratively, breaking it down into clear steps and working through them methodically.

\medskip
\textbf{Task Strategy}

\begin{enumerate}
  \item Analyze the user's request and set clear, achievable sub-goals.
  \item Start with a concise, numbered, step-by-step plan before taking any action.
  \item Work through these sub-goals sequentially and adjust the plan as needed.
  \item Use tools strategically to accomplish each sub-goal.
  \item Revise earlier steps if new information emerges.
\end{enumerate}

\medskip
\textbf{Tool-Use Guidelines}

\begin{enumerate}
  \item Each step must involve a single tool call, unless the task is already solved.
  \item Before each tool call:
  \begin{itemize}
    \item Summarize what is known.
    \item Identify what is missing.
    \item Choose the most relevant tool.
    \item Verify all required parameters.
  \end{itemize}
  \item All tool queries must include full context.
  \item Avoid vague queries. Each call should retrieve actionable information.
  \item Extract and summarize partial information if a tool result is incomplete.
\end{enumerate}

\medskip
\textbf{Tool-Use Communication Rules}

\begin{enumerate}
  \item Do not include tool results in your response.
  \item Do not present the final answer until the entire task is complete.
  \item Do not mention tool names.
  \item Do not engage in unnecessary back-and-forth.
  \item Do not use non-existent tools.
  \item Respond in the same language as the user's message.
  \item If the task does not require tool use, answer directly.
\end{enumerate}

\medskip
\textbf{Agent Specific Objective}

You are an agent that performs the task of searching and browsing the web for specific information and generating the desired answer. Your task is to retrieve reliable, factual, and verifiable information that fills in knowledge gaps.

Do not infer, speculate, summarize broadly, or attempt to fill in missing parts yourself. Only return factual content.

Critically assess the reliability of all information:
\begin{itemize}
  \item If the credibility of a source is uncertain, clearly flag it.
  \item Do not treat information as trustworthy just because it appears.
  \item If you find conflicting or ambiguous information, include all relevant findings and flag the inconsistency.
\end{itemize}

Be cautious and transparent in your output:
\begin{itemize}
  \item Always return all related information, even if incomplete, and flag any uncertainty.
  \item Never assume or guess. If an exact answer cannot be found, say so clearly.
  \item Prefer quoting or excerpting original source text rather than interpreting or rewriting it.
  \item Provide the URL of each source when available.
  \item If more context is needed, return a clarification request and do not proceed with tool use.
\end{itemize}

\end{promptbox}
\normalsize

\section{Disclosure of LLM Usage}
The LLM was used exclusively during editing (e.g., grammar, spelling, word choice). It plays no role in the ideation, research methodology, experimental design, or data analysis.
Authors are fully accountable for the manuscript, including any text generated or refined by the LLM, to ensure compliance with ethical guidelines and prevent plagiarism.

\end{document}